\documentclass{article}

\usepackage{PRIMEarxiv}

\usepackage[utf8]{inputenc} 
\usepackage[T1]{fontenc}    
\usepackage{hyperref}       
\usepackage{url}            
\usepackage{booktabs}       
\usepackage{amsfonts}       
\usepackage{nicefrac}       
\usepackage{microtype}      
\usepackage{lipsum}
\usepackage{fancyhdr}       
\usepackage{graphicx}       
\usepackage{multirow}
\usepackage{amsmath}
\usepackage{algorithm}
\usepackage{algorithmic}
\graphicspath{{media/}}     

\title{Meta-Transfer Derm-Diagnosis: Exploring Few-Shot Learning and Transfer Learning for Skin Disease Classification in Long-Tail Distribution
\thanks{This article has been accepted for publication in IEEE Journal of Biomedical and Health Informatics. ©2025 IEEE. Personal use of this material is permitted. Permission from IEEE must be obtained for all other uses, in any current or future media, including reprinting/republishing this material for advertising or promotional purposes, creating new collective works, for resale or redistribution to servers or lists, or reuse of any copyrighted component of this work in other works.
DOI:10.1109/JBHI.2025.3615479}
}

\author{
  Zeynep Özdemir* \\
  Department of Computer Engineering \\
  Ankara University, Ankara, Turkey \\
  \texttt{zynpozdemir@ankara.edu.tr} \\
  \And
  Hacer Yalim Keles \\
  Department of Computer Engineering \\
  Hacettepe University, Ankara, Turkey \\
  \texttt{hacerkeles@cs.hacettepe.edu.tr} \\
  \And
  Ömer Özgür Tanrıöver \\
  Department of Computer Engineering \\
  Ankara University, Ankara, Turkey \\
  \texttt{tanriover@ankara.edu.tr} \\
}

\begin{document}
\maketitle

\begin{abstract}
Building accurate models for rare skin diseases remains challenging due to the lack of sufficient labeled data and the inherently long-tailed distribution of available samples. These issues are further complicated by inconsistencies in how datasets are collected and their varying objectives. To address these challenges, we compare three learning strategies: episodic learning, supervised transfer learning, and contrastive self-supervised pretraining, within a few-shot learning framework. We evaluate five training setups on three benchmark datasets: ISIC2018, Derm7pt, and SD-198. Our findings show that traditional transfer learning approaches, particularly those based on MobileNetV2 and Vision Transformer (ViT) architectures, consistently outperform episodic and self-supervised methods as the number of training examples increases. When combined with batch-level data augmentation techniques such as MixUp, CutMix, and ResizeMix, these models achieve state-of-the-art performance on the SD-198 and Derm7pt datasets, and deliver highly competitive results on ISIC2018. All the source codes related to this work will be made publicly available soon at the provided \href{https://github.com/zeynepozdemir/MetaTransfer}{URL}.
\end{abstract}


\section{Introduction}
\label{sec:introduction}
\begin{figure}[h]
	\centering
	\includegraphics[width=0.5\textwidth]{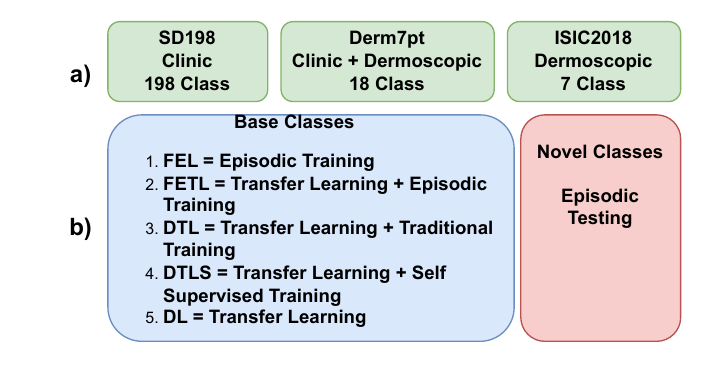}	
    \caption{Skin lesion classification framework: (a) benchmark datasets, (b) transfer learning strategies.}
	\label{fig_summary_diagram}
\end{figure}

Over the past decade, the field of medical image analysis has witnessed remarkable advancements, primarily driven by the development of deep convolutional neural networks and the availability of extensive labeled image datasets. These advancements have notably impacted various tasks, including organ segmentation \cite{organ_seg_chen2023magicnet,organ_seg_wang2023self}, tumor segmentation \cite{tumor_seg_hu2023label}, and disease detection \cite{zhou2023foundation}.
Although abundant data exists for common diseases, a significant gap persists in data availability for the over 6,000 known rare diseases, affecting approximately 7\% of the global population \cite{chung2021rare}. The diagnosis of these rare diseases, particularly certain skin conditions, presents unique challenges specific to the domain. Skin disease datasets, unlike those for natural image classification, often exhibit long-tailed distributions, where a few common classes dominate while rare classes are underrepresented. Additionally, variations in image quality and the presence of artifacts such as hair, rulers, and ink markings further complicate the classification process. The diversity of imaging modalities, including clinical and dermoscopic images, adds another layer of complexity, as these modalities vary in resolution, feature representation, and clinical relevance. Anatomical and biological diversity, subtle visual distinctions, and patient demographics introduce further complexities that directly impact model performance. Moreover, the need for expert clinicians to annotate datasets and strict data privacy regulations that limit data sharing represent significant obstacles in this field. These challenges underscore the necessity for developing advanced, domain-specific methodologies designed to address the complexities of skin disease classification \cite{liopyris2022artificial,hasan2023survey,yao2021single,mahajan2020meta}.

\begin{figure}[ht]
	\centering 
	\includegraphics[width=\textwidth]{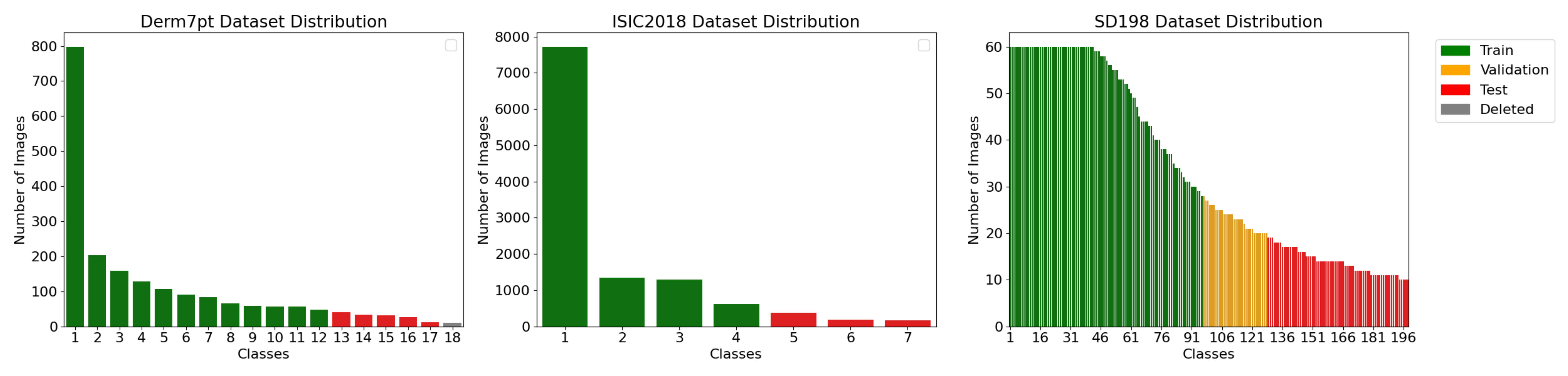}	
	\caption[width=\textwidth]{The figure shows the distributions of the datasets SD-198\cite{sd198dataset}, Derm7pt\cite{derm7ptdataset}, and ISIC2018\cite{isic2018dataset}, highlighting their long-tailed nature with some classes having very few instances. Base classes (common diseases) are marked as train (green) and validation (yellow), while novel classes (rare diseases) are labeled as test (red). In the Derm7pt dataset, classes with very few examples are shown as deleted (grey) and excluded from use.}
\label{fig_dist}
\end{figure}
 
Given these limitations, Few-shot learning (FSL) has emerged as a promising approach to address the challenges of class imbalance and data scarcity in medical imaging. FSL offers versatile solutions in image classification, detection, and segmentation tasks across both natural and medical images \cite{lang2023base,cheng2022holistic,lang2023retain,cheng2021spnet,quan2024dual,jiang2022multi,sun2023few,li2024tkr,he2021few}. Specifically, FSL methods have demonstrated notable effectiveness in navigating the complexities of skin disease classification \cite{mahajan2020meta}.

Building upon these advancements, various studies have been conducted to address the problem of skin disease classification using deep learning approaches. Recent advancements in this field are mainly in three categories: methods based on transfer learning \cite{hasan2022dermoexpert, jain2021deep}, those relying on few-shot learning \cite{prabhu2019few,li2020difficulty,mahajan2020meta,zhang2020st,zhu2020alleviating,singh2021metamed,li2023dynamic}, and approaches using cross-domain few-shot learning \cite{guo2020broader}. The state of the art models in this domain, such as Meta-DermDiagnosis, MetaMed, and PCN models \cite{mahajan2020meta,singh2021metamed,prabhu2019few} are designed to extract and learn high-level, domain-specific features during their training process.

In their study on the ISIC2018 dataset, Li et al. (2020) introduced the Difficulty Aware Meta-Learning (DAML) model, addressing task variability challenges \cite{li2020difficulty}. Similarly, Mahajan et al. (2020) developed the Meta-DermDiagnosis model, experimenting with the SD-198, Derm7pt, and ISIC2018 datasets \cite{mahajan2020meta}. This model incorporated Prototypical Networks and Reptile to enhance robustness but faced criticism for its reliance on symmetric dataset orientations. Singh et al. (2021) highlighted the efficacy of MixUp, CutOut, and CutMix as data augmentation techniques in the medical field when integrated with the MAML algorithm in their MetaMed model, particularly on the ISIC2018 dataset \cite{singh2021metamed}. Both MetaMed and Meta-DermDiagnosis models aim to enhance feature representation by broadening data augmentation and diversification. 

The mentioned studies in the field of skin disease classification have employed episodic learning as a method to acquire knowledge transferable to new classes. This approach involves dividing learning problems into small training and validation subsets to simulate scenarios encountered during evaluation. However, some studies in the FSL domain working with natural images have criticized episodic learning, arguing that its constraints are unnecessary and that using training groups in this manner is data-inefficient \cite{laenen2021episodes}. Similarly, \cite{tian2020rethinking} argued that, in tasks involving rare classes, the effectiveness of rapid adaptation depends more on the quality of the learned representation than on the few-shot learning algorithm itself. To enhance representation quality, they incorporated self-supervised learning components into their models. In a similar context, \cite{sun2019meta} proposed the concept of Meta-Transfer Learning, which aims to refine the learned representation. These studies conducted various experiments comparing the effectiveness of fine-tuning pre-trained Deep Neural Network (DNN) models with and without episodic learning. Their findings highlight the significance of integrating transfer learning with few-shot learning methodologies.

The existing literature consistently highlights the ongoing challenge of effectively managing the issue of long-tailed data distribution in skin disease studies \cite{mahajan2020meta, xiao2023boosting, dai2023pfemed}. Although current methods show promise in specific dataset contexts, their ability to generalize broadly remains limited. Most research in this field has focused on developing dataset-specific solutions, addressing the unique challenges and sensitivities arising from variations in class counts, data formats, and other critical characteristics. On the other hand, episodic training techniques, which have been criticized for their inefficiency in the natural image domain of FSL, continue to be utilized in this context.

In this study, we aim to explore  the effects of episodic learning, transfer learning and self-supervised learning for rare skin disease classification. One of our objective was to  evaluate different model training approaches, using a benchmark test set specifically for rare skin diseases.  In addition to transfer learning approaches using supervised ImageNet-pretrained models such as ResNet50 \cite{resnet_targ2016}, DenseNet121 \cite{huang2017densely}, MobileNetV2 \cite{sandler2018mobilenetv2}, Vision Transformer (ViT) \cite{vit_dosovitskiy2020image}, ConvNeXt \cite{convnext_liu2022convnet}, and EfficientNet \cite{efficientnet_tan2019}, we extended our evaluation to include models pretrained via contrastive self-supervised learning, including SimCLR \cite{simclr_chen2020simple}, MoCo-v3 \cite{mocov3_chen2021empirical}, DINO-v2 \cite{dinov2_oquab2023}, SimSiam \cite{simsiam_chen2021exploring}, SwAV \cite{swav_caron2020unsupervised}, and ViT-MMF \cite{vit_mmf_liu2023improving}. Additionally, we investigated the potential of transformer-based foundation models developed for medical imaging, such as BioViL \cite{biovil_boecking2022making} and MedCLIP \cite{medclip_wang2022}.

In this context, we implemented five distinct training approaches, each evaluated using consistent metrics for novel classes (Figure \ref{fig_summary_diagram}). Our first approach, Few-Shot Episodic Learning (FEL), applies episodic training to models initialized with random weights (i.e., without any pre-training). The second, Few-Shot Episodic Transfer Learning (FETL), extends this by incorporating ImageNet-pretrained weights, which are fine-tuned on the long-tail skin disease dataset within an episodic learning setup. The third method, Deep Transfer Learning (DTL), omits episodic training and instead fine-tunes ImageNet-pretrained models directly on base classes. The fourth, Deep Transfer Learning with Self-Supervised Learning (DTLS), uses contrastively pretrained models, which are then fine-tuned with supervised training on the target dataset. Finally, Deep Learning (DL) serves as a baseline, where ImageNet-pretrained models are evaluated without any additional training or fine-tuning.

FETL, DTL, and DTLS were designed to incorprate FSL natural image domain advancements to address rare challenge. Additionally, performance improvements were targeted by integrating data augmentation techniques, such as MixUp, CutMix, and ResizeMix, specifically adapted for the unique properties of skin disease datasets.

To evaluate the effectiveness of explored methods we carried out extensive testing across three benchmark skin disease datasets: SD-198, Derm7pt, and ISIC2018. Our comprehensive analysis provide insights for effective strategies in tackling the challenge of long-tailed data distribution in skin diseases. The key contributions of our study are summarized as follows:

\begin{itemize}

    \item We  evaluate different model training approaches, using a  benchmark test set specifically curated for long-tail distributions in rare skin diseases. The final comparison is conducted through episodic testing. To the best of our knowledge, this is the first time such a thorough methodological analysis has been conducted in this domain.rare skin disease.

    \item By comparing episodic and traditional training methods, our findings indicate that traditional training becomes increasingly beneficial as the number of shots (training examples) grows.

    \item We systematically evaluated a wide range of pretrained and foundation models (e.g., ConvNeXt, ViT, ResNet50, MedCLIP, BioVIL) under few-shot settings, both with and without additional fine-tuning. Our results provide valuable insights into selecting strong initialization points for rare skin disease classification, emphasizing the importance of pretraining choices in low-data regimes.
    
    \item  We demonstrated that combining transfer learning and self-supervised learning with few-shot learning significantly enhances both the learned representation and the testing performance in the context of rare skin diseases.
\end{itemize} 

\section{Related Works}

\subsection{Transfer Learning for Medical Image Domain}

In deep learning, transfer learning involves adapting a pre-trained model for a new task, commonly referred to as fine-tuning (FT). Models pre-trained on large datasets have shown superior generalization compared to randomly initialized ones \cite{erhan2010does}. Various techniques facilitate the transfer of knowledge between source and target domains \cite{he2017mask, huang2017speed, ying2018transfer, zamir2018taskonomy}.

For skin disease classification, transfer learning is widely used. Architectures such as EfficientNet, ResNet, and DenseNet, typically pre-trained on ImageNet or dermatological datasets, have been transferred to datasets like ISIC2017 and ISIC2018 \cite{mahbod2020transfer, qin2020gan, liu2020automatic, mahbod2019fusing}, demonstrating notable performance gains in data-scarce scenarios. A more comprehensive discussion is provided in \cite{atasever2023comprehensive}.

Beyond conventional supervised pretraining, recent advances in self-supervised learning (SSL) have introduced novel mechanisms for learning transferable and robust representations without labels. In few-shot skin disease classification, studies such as MetaMed \cite{singh2021metamed} explored episodic training with data augmentation, using shallow architectures and transfer learning on ISIC2018. In contrast, our approach leverages deeper architectures to enhance feature representation, aiming to better capture the complexity of rare skin disease patterns.

\subsection{Few-Shot Learning in Computer Vision}

Few-shot learning (FSL) aims to recognize novel classes with only a few labeled examples, leveraging a substantial number of examples from base classes. FSL algorithms can be broadly categorized into three groups: initialization-based methods, metric learning-based methods, and hallucination-based methods.

In this context, initialization-based methods take a \textit{learning to fine-tune} approach. They aim to acquire an effective model initialization, specifically the neural network parameters. This facilitates the adaptation of classifiers with limited labeled examples through a few gradient update steps for new classes \cite{finn2017model, nichol2018reptile, rusu2018meta}. Another strategy involves distance metric learning methods, embracing a \textit{learning to compare} paradigm for few-shot classification. These methods are foundational approaches utilizing encoded feature vectors and a distance measurement metric based on the nearest-neighbor principle to assign labels. For instance, Prototypical Networks \cite{snell2017prototypical} utilizes Euclidean distance, Matching Networks \cite{vinyals2016matching} employs cosine similarity, and Relation Networks \cite{sung2018learning} utilizes its own CNN-based measurement module for this purpose \cite{li2023deep}. Additionally, hallucination-based methods directly address data scarcity through \textit{learning to augment}. Here, hallucination involves generating data not derived from real examples or direct observations. The generator's objective is to transfer appearance variations present in the base classes to novel classes \cite{hariharan2017low, antoniou2017data}. 

The mentioned FSL methods adopt an episodic training approach during the training of the data (base classes). However, some studies have shown that training the data by dividing it into tasks leads to inefficient use of the available data \cite{laenen2021episodes, tian2020rethinking}. Additionally, \cite{kumar2023effect} has demonstrated, contrary to the prevailing notion, through experiments conducted with benchmark datasets and fundamental FSL algorithms, that an increase in task diversity, proportional to the increase in classes and data, does not lead to an improvement in success. Therefore, we can categorize FSL approaches into two groups based on their training processes: meta-learning-based and transfer-learning (TL)-based methods.
Among TL methods, S2M2-R \cite{mangla2020charting}, Baseline \cite{chen2019closer}, PT-MAP \cite{hu2021leveraging}, and Meta-Transfer Learning \cite{sun2019meta} train on base classes using a standard classification network and fine-tune the classifier head on episodes generated from new classes. These methods aim to train a powerful feature extractor that produces transferable features for the new class. Experimental methods have demonstrated that these approaches can achieve more effective and higher performance compared to previous FSL methods, utilizing a simpler and more efficient process. Due to the superior performance of TL-based methods, we explored this approach in conjunction with Prototypical Networks as a way to predict rare skin diseases.

\subsection{ Few-Shot Learning for Skin Disease Classification}

The imbalanced distribution of skin disease classes and factors such as limited image availability in rare diseases necessitate the application of few-shot learning methods. As a solution to this challenge, \cite{li2020difficulty} proposed the Difficulty-Aware Meta-Learning (DAML) model based on Meta-Learning. This model adjusts the losses for each task, increasing the weight of challenging tasks while decreasing that of easier tasks, thereby emphasizing and increasing the importance of difficult tasks. This study, aiming to highlight more distinctive features for each class, can be categorized as initialization-based FSL. On the other hand, the model named MetaMed, evaluated in both initialization and hallucination-based  FSL categories, by \cite{singh2021metamed}, combines advanced data augmentation techniques such as mixup, cutout, and cutmix with the reptile model during training to enhance its generalization capabilities. Working with the ISIC2018 dataset, they compared the transfer learning approach of their proposed model with other studies, reporting an average improvement of up to 3\% in performance. In the metric-based branch, several methods have been proposed. \cite{zhu2020alleviating} advocated for the superiority of the Query-Relative loss over the Cross-Entropy loss commonly used in FSL. Additionally, \cite{zhu2021temperature} suggested the utilization of Temperature Networks alongside Prototype Networks. They adapted specific temperatures for different categories to reduce intra-class variability and enhance inter-class dispersion. Moreover, they applied penalization based on the proximity of query examples. \cite{mahajan2020meta} introduced a model named Meta-DermDiagnosis, aiming to obtain invariant features after various transformations by replacing traditional convolutional layers with group-equivariant convolutions, similar to Prototypical Networks and Reptile.

In the context of transfer-learning-based algorithms, the study in\cite{dai2023pfemed} proposed a model named PFEMed, suggesting a dual-encoder structure. This brings about one encoder with fixed weights pre-trained on large-scale public image classification datasets and another encoder trained on the target medical dataset. On the other hand, \cite{xiao2023boosting} designed a dual-branch framework and improved performance using a model employing prototypical networks and contrastive loss. To address the challenge of observing diverse subgroups within dermatological disease clusters, \cite{li2023dynamic} introduced the Sub-Cluster-Aware Network (SCAN) model. SCAN utilizes a dual-branch structure to enhance feature explanation, learning both class-specific features for disease differentiation and subgroup-related features.

Building upon previous studies, we present a comprehensive framework for few-shot rare skin disease classification by systematically evaluating a wide range of backbone architectures and training strategies. Specifically, we experimented with architectures including ResNet50 \cite{resnet_targ2016}, DenseNet121 \cite{huang2017densely}, MobileNetV2 \cite{sandler2018mobilenetv2}, ViT \cite{vit_dosovitskiy2020image}, ConvNeXt \cite{convnext_liu2022convnet}, and EfficientNet \cite{efficientnet_tan2019}, most of which were pretrained using supervised learning on ImageNet. To explore alternative pretraining paradigms, we also employed contrastive self-supervised methods such as SimCLR \cite{simclr_chen2020simple}, MoCo-v3 \cite{mocov3_chen2021empirical}, SwAV \cite{swav_caron2020unsupervised}, and SimSiam \cite{simsiam_chen2021exploring}, as well as distillation-based approaches like DINOv2 \cite{dinov2_oquab2023} and ViT-MMF \cite{vit_mmf_liu2023improving}, particularly on ResNet50 and ViT backbones. Additionally, transformer-based foundation models tailored for medical imaging, including BioViL \cite{biovil_boecking2022making} and MedCLIP \cite{medclip_wang2022}, were included as part of our extended transfer learning evaluation.

Our comparative analysis focuses on three major training paradigms: traditional supervised transfer learning, contrastive self-supervised pretraining, and episodic few-shot learning. While contrastive pretraining showed benefits in certain cases, our results indicate that supervised transfer learning generally yields more robust performance in few-shot settings. Furthermore, episodic learning methods often failed to fully exploit the available data. To improve generalization, we incorporated augmentation techniques such as MixUp, CutMix, and ResizeMix, customized to address the characteristics of dermatological datasets. Importantly, our evaluations rely solely on existing benchmark datasets and do not require any supplementary external data, making the findings potentially applicable to other low-resource rare disease classification tasks. Detailed results and comparative insights are presented in Section~\ref{sec:SOTA}.

\section{Datasets and Evaluation}
\label{sec:Datasets}

\begin{figure}[h]
	\centering
	\includegraphics[width=0.8\textwidth]{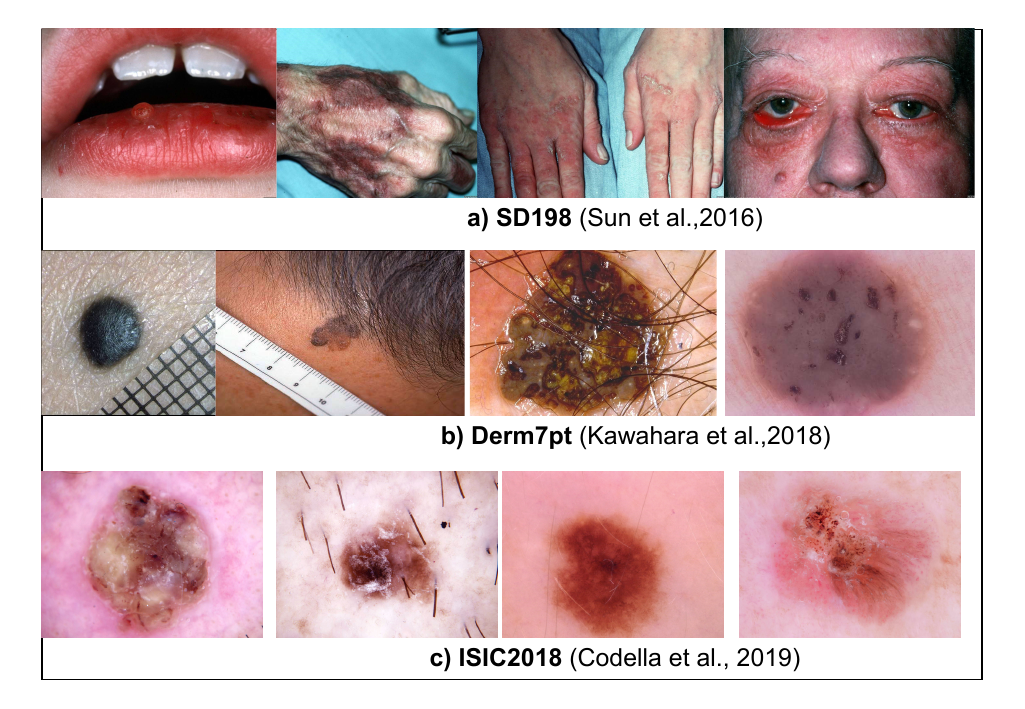}	
	\caption{Some sample images from skin disease classification datasets} 
	\label{fig_dataset}
\end{figure}

All datasets used in this study were standardized to an input size of 224 $\times$ 224 pixels for consistency with prior works \cite{mahajan2020meta, li2023dynamic, dai2023pfemed}. Their distributions are presented in Figure~\ref{fig_dist}, while representative examples are shown in Figure~\ref{fig_dataset}.  

\textbf{The SD-198 dataset} \cite{sd198dataset} comprises 6,584 clinical images across 198 categories (e.g., eczema, acne, rosacea, and skin cancers). Captured using digital cameras or mobile phones at 1640 $\times$ 1130 resolution, the images exhibit broad diversity in patient age, gender, skin type, disease stage, and imaging conditions. Following previous works, 70 rare classes (fewer than 20 samples) were designated for testing, while the remaining 128 classes served for training.  

\textbf{The Derm7pt dataset} \cite{derm7ptdataset} contains over 2,000 clinical and dermoscopic images grouped into 20 classes, annotated using a 7-point checklist for malignancy. Two categories \textit{miscellaneous} (unspecified diseases) and \textit{melanoma} (single case) were excluded, leaving 18 categories, of which 13 were used for training and 5 for testing (10–34 samples per novel class).  

\textbf{The ISIC 2018 dataset} \cite{isic2018dataset} includes 10,015 dermoscopic images across seven lesion classes labeled by expert pathologists. Following the standard split, 7,515 images form the training set and 2,500 form the test set. In line with prior studies, a subset comprising four base classes and three novel classes was selected for few-shot classification tasks.

\begin{figure*}[ht!]
	\centering 
	\includegraphics[width=\textwidth]{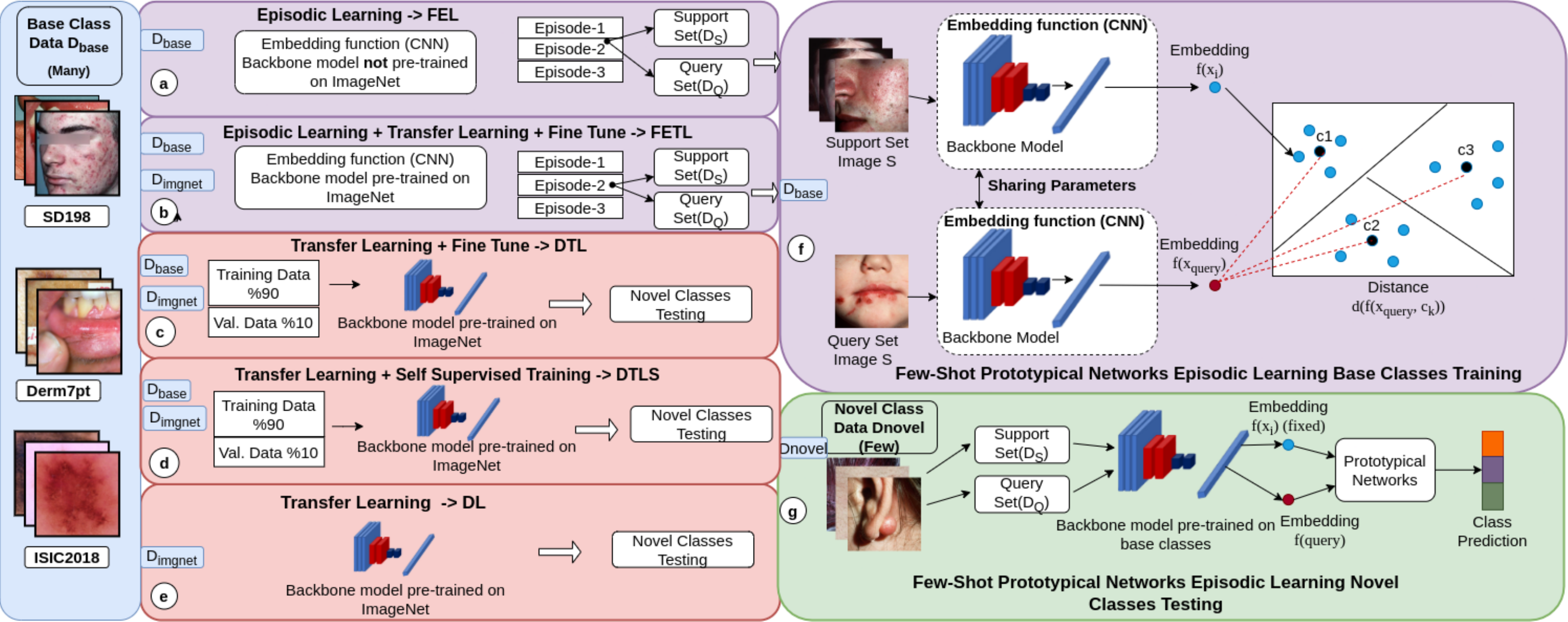}	
	\caption[width=\textwidth]{Overall framework of our pipeline: Meta-Transfer Derm-Diagnosis. a) Episodic learning is combined with DenseNet and MobileNet architectures without the use of ImageNet weights. b) ImageNet pre-trained weights are utilized along with the application of an episodic learning strategy.  c) Pre-trained weights and all base class data are employed for fine-tuning with ImageNet. d) Pre-trained weights and all base class data are employed for self-supervised training with ImageNet. e) Only ImageNet weights are utilized without fine-tuning. e) Detailed diagram illustrating the use of episodic learning and prototypical networks. It is applied in the continuation of parts a and b. g) Common evaluation scheme using novel data across segments a, b, c, d and e.} 

\label{fig_framework}
\end{figure*}

\begin{figure}[h]
	\centering
	\includegraphics[width=0.7\textwidth]{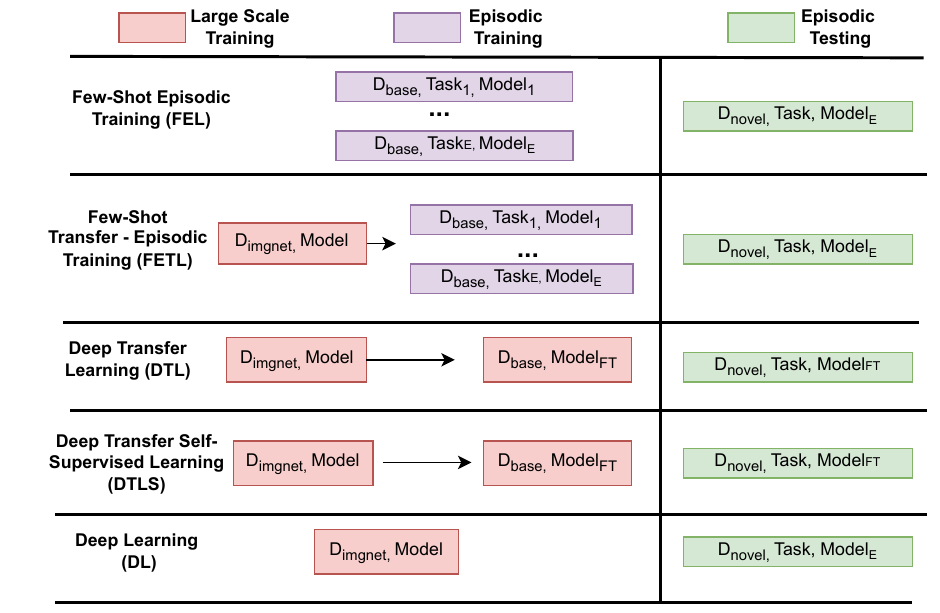}	
	\caption{The flowchart illustrating the components and the training strategies of the FEL, FETL, DTL, DTLS and DL models within the proposed framework.} 
	\label{fig_sema}
\end{figure}

\section{The Methodology}

This section outlines the core methodologies deployed in our study, starting with the Meta-Transfer Derm-Diagnosis Framework. This framework is pivotal for assessing five distinct model training strategies that are particularly valuable in scenarios with limited data samples. These strategies include Few-Shot Episodic Transfer Learning (FETL), Few-Shot Episodic Learning (FEL), Deep Transfer Learning (DTL), Deep Transfer Learning with Self-Supervised Adaptation (DTLS), and Standard Deep Learning (DL).

We implement a precise evaluation process, employing a combination of the selected benchmark datasets (Section \ref{sec:Datasets}), and a specialized testing approach. This ensures a comprehensive and fair analysis of each training method. Subsequently, we provide an overview of Prototypical Networks and Transfer Learning. 

\subsection{Meta-Transfer Derm-Diagnosis Framework}

The proposed Meta-Transfer Derm-Diagnosis Framework is used to effectively combine few-shot training methods with transfer learning. This integration is specifically designed to improve model performance for tail classes, which often have limited data, thereby leveraging the complementary strengths of both methodologies.

Few-shot classification operates with two distinct datasets: the base dataset, denoted as $D_{\text{base}}$, and the novel dataset, denoted as $D_{\text{novel}}$. The novel dataset, $D_{\text{novel}}$, is utilized for the actual classification task, while the base dataset, $D_{\text{base}}$, helps in training the classifier by transferring essential knowledge from it. Additionally, during training, ImageNet dataset  is also used and will be referred to as the domain $D_{\text{imgnet}}$. For clarity and coherence, let us include two definitions adapted to our framework from prior literature \cite{li2023deep} that are related to our study's context in few-shot learning. These definitions will help in framing our approach and methodology.

\textbf{Definition 1.} For the training and testing phases, the novel dataset $D_{\text{novel}}$ is split into two subsets: the support set ($D_S$) and the query set ($D_Q$). In a typical \textbf{few-shot classification} scenario, the support set $D_S$ contains only a few samples per class, often ranging from 1 to 5. The primary goal in few-shot classification is to train a classifier, $f: X_{\text{novel}} \rightarrow Y_{\text{novel}}$, using the limited data in $D_S$. The classifier should then be able to accurately categorize the instances in the query set $D_Q$. If $D_S$ includes $N$ distinct classes with $K$ labeled examples per class, this scenario is defined as an $N$-way $K$-shot classification. The case with only one labeled example per class is termed one-shot classification.

\textbf{Definition 2.} A few-shot classification task is referred to as \textit{cross-domain few-shot classification} when the train dataset $D_{\text{train}}$ and the novel dataset $D_{\text{novel}}$ are sourced from distinct domains. 

To further clarify our methodology, we define the datasets used. The base dataset is defined as 

$ D_{\text{base}} = \{ (x_i, y_i) ;  x_i \in X_{\text{base}}, y_i \in Y_{\text{base}} \}_{i=1}^{N_{\text{base}}}$, where $x_i$ represents the feature vector of the $i$-th image, and $y_i$ is its corresponding class label. The novel dataset is similarly represented as 

$D_{\text{novel}} = \{ (\tilde{x}_j, \tilde{y}_j) ; \tilde{x}_j \in X_{\text{novel}}, \tilde{y}_j \in Y_{\text{novel}} \}_{j=1}^{N_{\text{novel}}}$. It is crucial to note that the class labels in $D_{\text{base}}$ and $D_{\text{novel}}$ are mutually exclusive, i.e., $Y_{\text{base}} \cap Y_{\text{novel}} = \emptyset$. In a similar manner, $D_{\text{imgnet}}$ is represented as:  $\{ (\bar{x}_k, \bar{y}_k) ; \bar{x}_k \in X_{\text{imgnet}}, \bar{y}_k \in Y_{\text{imgnet}} \}_{k=1}^{N_{\text{imgnet}}}$. 

To rigorously evaluate the framework we devised, detailed in Figure \ref{fig_framework}, we maintained a consistent evaluation by keeping  $D_{\text{novel}}$ fixed. Here, $D_{\text{train}}$ denotes the datasets used during training. We formulated five distinct training methodologies, each independently designed: FETL, where $D_{\text{train}} = D_{\text{imagenet}} + D_{\text{base}}*$ ; FEL, where $D_{\text{train}} = D_{\text{base}}*$; DTL and DTLS, where $D_{\text{train}} = D_{\text{imagenet}} + D_{\text{base}}$; and DL, where $D_{\text{train}} = D_{\text{imagenet}}$. The notation $D_{\text{base}}*$ indicates episodic training, while the others involve traditional large-scale training approaches for the corresponding datasets in that domain.

In accordance with these definitions, as summarized in Figure \ref{fig_sema}, our FETL, FEL, DTLS and DTL models utilize $D_{\text{base}}$ and $D_{\text{novel}}$ classes from the same domain for training and testing as specified in Definition 1. On the other hand, the proposed DL model utilizes two separate domains for training and testing; hence it aligns with Definition 2. Therefore, three out of our five proposed analyses include adapted few-shot training and testing methodologies considering the data extracted carefully from the same long-tail distributions (FETL, FEL, DTL and DTLS), while the last one (DL) corresponds to a cross-domain evaluation.

\begin{algorithm}
\caption{$N$-Way $K$-Shot Classification Evaluation.}
\label{algorithm1}
\begin{algorithmic}[1]
\REQUIRE $ D_{\text{train}} = \{ (x_i, y_i) ;  X_i \in \mathcal{X}_{\text{train}}, Y_i \in \mathcal{Y}_{\text{train}} \}_{i=1}^{N_{\text{train}}}.$
\REQUIRE $ D_{\text{novel}} = \{ (\tilde{x}_j, \tilde{y}_j) ;  \tilde{x}_j \in \mathcal{X}_{\text{novel}}, \tilde{y}_j \in \mathcal{Y}_{\text{novel}} \}_{j=1}^{N_{\text{novel}}}. $
\REQUIRE Number of episodes $E$
\FOR{$e = 1,...,E$}
  \STATE Randomly select $N$ classes from $\mathcal{Y}_{\text{novel}}$.
  \STATE Randomly select $K$ samples from each class as the support set $D_{\text{S}}^{(e)}$.
  \STATE Randomly select $M$ samples from the remaining samples of $N$ classes as the query set $\{(\tilde{x}^{(e)},\tilde{y}^{(e)})\}$.
  \STATE Record predicted labels $\hat{y}^{(e)} = f(\tilde{x}^{(e)} |  D_{\text{train}}, D_{\text{S}}^{(e)})$.
  \STATE Compute accuracy $a^{(e)} = \frac{1}{M} \sum_{m=1}^{M} 1 [\hat{y}^{(e)} = {\tilde{y}}^{(e)}]$
\ENDFOR
\STATE Compute: $Avg\_Acc = \frac{1}{E} \sum_{e=1}^{E} a^{(e)}$
\RETURN $Avg\_Acc$
\end{algorithmic}
\end{algorithm}

The evaluation of our classifier in $N$-way $K$-shot classification is outlined in Algorithm 1. This process includes a series of episodes, each presenting a unique classification task, allowing for a thorough assessment of the classifier's performance. In this procedure, as the initial step, we randomly select $N$ classes from the novel set. Following this, we randomly choose $K$ samples from each of these $N$ classes to constitute a support set. Concurrently, we select $M$ samples from the remaining instances in these classes to form a query set. For each episode, denoted as the $e$'th episode, the query set's instances and labels are represented as $\tilde{x}^{(e)}$ and $\tilde{y}^{(e)}$, respectively. A learning algorithm is then applied, utilizing the model that is pretrained with $D_{train}$ and tuned to the support set of the $e$'th episode, $D_S^{(e)}$. This algorithm yields a classifier that predicts labels for the instances in the query set. To quantify the classifier’s performance, we calculate the classification accuracy for each episode, referred to as $a^{(e)}$. The overall effectiveness of the learning algorithm is then determined by averaging these classification accuracies across all episodes.

\subsection{Prototypical Networks }

As depicted in Figure \ref{fig_framework} (top right section), we used Prototypical Networks \cite{snell2017prototypical} for both episodic training and testing. It is a meta-learning approach, which is designed to represent each class through a prototype vector, based on distance metrics. This vector is an average of embedded instances in a support set, specifically linked to that class. Formally, for a set of $N$ classes, the support set $S =\{ (x_1, y_1), ..., (x_N, y_N)\}$ is constructed, where each $x_i \in \mathbb{R}^D$ is a $D$-dimensional feature vector, and $y_i$ is its corresponding label in the range $\{1, . . . , K\}$.

Each class prototype $c_k \in \mathbb{R}^{M}$ is computed as the mean vector of its associated embedded instances, using an embedding function $f_{\phi} : \mathbb{R}^D \rightarrow \mathbb{R}^M$ with learnable parameters $\phi$. The prototype for class $k$ is derived as:

\begin{equation}\label{eq:eq_1}
c_k = \frac{1}{|S_k|} \sum_{(x_i, y_i) \in S_k} f_{\phi}(x_i).
\end{equation}

The network utilizes a distance function $d : \mathbb{R}^{M} \times \mathbb{R}^{M} \rightarrow [0, +\infty)$ to calculate the probability distribution across classes for a query point $x$, based on the softmax of distances between the query point and class prototypes:

\begin{equation}\label{eq:eq_2}
p_{\phi}(y=k |x) = \frac{exp(-d(f_{\phi}(x),c_k))}{\sum_{k'}{exp(-d(f_{\phi}(x),c_{k'}))}}.
\end{equation}

Training involves minimizing the negative log likelihood $J(\phi) = -log p_{\phi}(y = k | x)$ of the true class $k$, using Stochastic Gradient Descent (SGD). 

\subsection{Regularization via Image Augmentation }

Regularization is vital for preventing overfitting and improving the generalization of deep models. Among these, image augmentation is particularly effective in supervised learning. However, common techniques such as rotation or flipping are often insufficient in data-limited domains like medical imaging. To address this, advanced methods such as MixUp \cite{zhang2017mixup}, CutMix \cite{yun2019cutmix}, and ResizeMix \cite{qin2020resizemix} have been proposed, enabling the generation of more diverse samples and improving model robustness.

\section{Experiments and Discussion}

\begin{table*}[!]
\fontsize{8}{9}\selectfont 
\centering

\caption{Few-shot classification accuracies of backbone models without fine-tuning on \textbf{SD-198},  \textbf{Derm7pt}, and \textbf{ISIC2018} datasets.}

\begin{tabular}{llllllllllll}
\hline
\multirow{2}{*}{\textbf{Methods}} & \multicolumn{1}{c}{\multirow{2}{*}{\textbf{\begin{tabular}[c]{@{}c@{}}Params\\ (M)\end{tabular}}}} & \multicolumn{1}{c}{\multirow{2}{*}{\textbf{\begin{tabular}[c]{@{}c@{}}Flops\\ (G)\end{tabular}}}} & \multicolumn{3}{c}{\textbf{SD-198}} & \multicolumn{3}{c}{\textbf{Derm7pt}} & \multicolumn{3}{c}{\textbf{ISIC2018}} \\ \cline{4-12} 
 & \multicolumn{1}{c}{} & \multicolumn{1}{c}{} & \textbf{2W1S} & \textbf{2W5S} & \textbf{5W5S} & \textbf{2W1S} & \textbf{2W5S} & \textbf{5W5S} & \textbf{2W1S} & \textbf{2W5S} & \textbf{3W5S} \\ \hline
MedCLIP(ViT) \cite{medclip_wang2022}         & 27.94  & 4.5 & 59.28 & 64.42 & 36.6 & 57.88 & 67.71 & 31.16 & 50.74 & 51.76 & 35.09 \\
MedCLIP(ResNet50) \cite{medclip_wang2022}    & 25.56  & 4.11 & 66.58 & 80.56 & 60.06 & 57.66 & 64.62 & 38.25 & 51.96 & 55.64 & 39.5 \\ 
BioVIL(ResNet-ViT) \cite{biovil_boecking2022making}   & 25.13  & 4.2 & 61.51 & 72.25 & 47.96 & 56.46 & 62.52 & 36.8 & 50.95 & 52.82 & 36.49 \\ \hline
ConvNeXt(Base) \cite{convnext_liu2022convnet}         & 88.59  & 15.36  & \underline{80.32} & \underline{92.86} & 81.15 & 60.25 & 76.84 & 55.95 & 59.55 & 72.27 & 58.57 \\ 
EfficientNetB5 \cite{efficientnet_tan2019}         & 30.39  & 10.8    & 72.88 & 85.56 & 70.12 & 56.0 & 64.45 & 40.05 & \underline{62.14} & 69.91 & 54.35 \\ 
ResNet50 \cite{resnet_targ2016}                & 25.56  & 4.12    & 76.77 & 92.39 &  \underline{82.15} & 58.45 & 74.21 & 55.06 & 59.97 & 76.49 & 63.61 \\ 
EfficientNetB4 \cite{efficientnet_tan2019}         & 19.34  & 4.66    & 74.59 & 88.44 & 74.33 & 56.55 & 68.78 & 44.58 & \textbf{62.91} & 72.56 & 57.0 \\
DenseNet121 \cite{huang2017densely}            & 7.98   & 2.88     & 75.7 & 91.86 & 81.03 & 59.42 & 75.89 & 56.83 & 60.66 & 76.4 & 62.72 \\ 
MobileNetV2 \cite{sandler2018mobilenetv2}            & 3.5    & 0.32      & 77.77 & \textbf{92.96} & \textbf{83.24} & 58.65 & 74.15 & 54.69 & 60.55 & 76.42 & 63.22 \\\hline
MoCo-v3(ResNet50)  \cite{mocov3_chen2021empirical}  & 68.01  & 4.11 & 69.06 & 89.52 & 78.08 & 58.82 & 76.53 & 58.43 & 58.6 & 76.58 & 64.52 \\ 
SimSiam(ResNet50) \cite{simsiam_chen2021exploring} & 38.2   & 4.11 & 71.28 & 90.35 & 79.11 & 58.81 & 76.14 & 58.57 & 60.55 & \textbf{78.05} & \textbf{65.74} \\ 
SwAV(ResNet50) \cite{swav_caron2020unsupervised}     & 28.35  & 4.11 & 70.24 & 89.57 & 77.8 & 57.69 & 74.95 & 56.64 & 58.51 & \underline{76.88} & \underline{64.88} \\ 
SimCLR(ResNet50) \cite{simclr_chen2020simple}   & 27.97  & 4.11 & 71.55 & 89.9 & 77.93 & 58.26 & 74.47 & 56.34 & 59.14 & 75.11 & 62.14  \\ \hline
MoCo-v3(ViT)  \cite{mocov3_chen2021empirical}           & 215.68 & 17.58 & 75.58 & 85.77 & 70.08 & \textbf{88.4} & \underline{92.18} & 65.24 & 53.13 & 55.98 & 40.18 \\
DINOv2(ViT) \cite{dinov2_oquab2023} & 86.58  & 152.0 & 70.44 & 79.03 & 57.63 & \underline{86.89} & 90.07 & 57.97 & 54.02 & 58.04 & 42.21 \\ 
ViT-MMF \cite{vit_mmf_liu2023improving}              & 86.57  & 17.58 & 71.69 & 87.65 & 73.74 & 81.92 & 91.05 & \underline{75.62} & 53.37 & 58.93 & 43.46 \\
 \hline
ViT(Base) \cite{vit_dosovitskiy2020image}        & 86.57  & 17.58 & \textbf{81.34} & 92.55 & 82.14 & 86.85 & \textbf{93.05} & \textbf{77.9} & 52.87 & 58.28 & 42.31 \\ \hline
\end{tabular}

\label{tab:Models_Backboned}
\end{table*}

\begin{table*}[]
\fontsize{9}{11}\selectfont
\centering
\caption{Few-shot classification accuracies of fine-tuned models on \textbf{SD-198},  \textbf{Derm7pt}, and \textbf{ISIC2018} datasets.}

\resizebox{\textwidth}{!}{ 
\begin{tabular}{l|lllll|lllll|lllll}
\hline
\multirow{2}{*}{\textbf{Methods}} & \multicolumn{5}{c|}{\textbf{SD-198}} & \multicolumn{5}{c|}{\textbf{Derm7pt}} & \multicolumn{5}{c}{\textbf{ISIC2018}} \\ \cline{2-16} 
 & \textbf{2W1S} & \textbf{2W5S} & \textbf{2W10S} & \textbf{5W1S} & \textbf{5W5S} & \textbf{2W1S} & \textbf{2W5S} & \textbf{2W10S} & \textbf{5W1S} & \textbf{5W5S} & \textbf{2W1S} & \textbf{2W5S} & \textbf{2W10S} & \textbf{3W1S} & \textbf{3W5S} \\ \hline

ResNet50+Aug DTL & 78.53 & 94.47 & \underline{97.52} & 59.52 & 87.24 & 61.62 & 78.86 & 86.71 & 34.03 & 60.41 & 61.7 & 78.28 & 83.47 & 46.72 & 65.86 \\
ResNet50(SimCLR)+Aug DTLS & 82.56 & 93.00 & 94.97 & 62.01 & 79.20 & \underline{63.91} & \underline{80.58} & 85.62 & \underline{36.52} & \underline{63.26} & 62.9 & 78.54 & 82.88 & 47.58 & 66.1 \\
ResNet50(SimSiam)+Aug DTLS & 71.28 & 90.35 & 95.08 & 47.09 & 76.01 & 58.81 & 76.14 & 83.13 & 31.19 & 58.57 & 60.65 & 78.05 & 82.40 & 45.05 & 65.74 \\
ResNet50(MoCo-v3)+Aug DTLS & 69.09 & 89.52 & 94.88 & 43.92 & 76.00 & 58.82 & 76.53 & 83.36 & 31.15 & 58.43 & 58.6 & 76.58 & 81.04 & 43.32 & 64.52 \\ \hline

DenseNet121 FEL & 82.91 & 92.95 & 95.69 & 63.61 & 82.05 & 61.4 & 72.46 & 75.68 & 31.78 & 47.05 & 57.86 & 65.97 & 68.58 & 42.08 & 50.39 \\
DenseNet121 FETL & 82.83 & 93.58 & 95.05 & 64.81 & 84.02 & 60.99 & 74.12 & 78.28 & 33.74 & 52.48 & 59.04 & 67.51 & 70.07 & 42.25 & 51.56 \\
DenseNet121 DTL & 81.21 & 94.56 & 95.74 & 62.61 & 86.95 & 60.0 & 77.25 & 83.9 & 32.04 & 57.68 & 58.55 & 76.69 & 81.99 & 43.63 & 64.47 \\
DenseNet121+Aug DTL & 82.29 & \textbf{95.46} & \textbf{97.97} & 65.54 & \textbf{88.76} & 62.56 & 81.00 & \underline{87.16} & 34.75 & 61.91 & 56.67 & 65.06 & 67.85 & 39.81 & 48.66 \\ \hline

MobileNetV2 FEL & 82.66 & 91.49 & 93.19 & 61.73 & 78.16 & 59.37 & 69.18 & 71.98 & 32.38 & 45.38 & 58.42 & 66.94 & 69.5 & 42.33 & 50.41 \\
MobileNetV2 FETL & \textbf{84.77} & 93.66 & 95.10 & \underline{65.85} & 83.22 & 61.38 & 72.34 & 76.38 & 32.79 & 47.81 & 58.49 & 66.31 & 68.82 & 43.09 & 50.92 \\
MobileNetV2 DTL & 82.42 & 94.4 & 96.77 & 64.06 & 86.67 & 59.65 & 75.41 & 82.43 & 31.8 & 55.92 & \underline{64.83} & \underline{81.63} & \underline{85.18} & \underline{49.51} & \underline{69.35} \\
MobileNetV2+Aug DTL & \underline{84.63} & \textbf{95.46} & 97.34 & \textbf{67.94} & \underline{88.49} & 60.37 & 77.41 & 84.93 & 33.54 & 59.45 & \textbf{65.55} & \textbf{83.21} & \textbf{86.83} & \textbf{50.40} & \textbf{71.28} \\ \hline
ViT(Base)+LoRA+Aug DTL & 82.04 & 92.80 & 95.80 & 64.76 & 85.60 & \textbf{86.35} & \textbf{93.68} & \textbf{95.56} & \textbf{66.33} & \textbf{80.74} & 61.59 & 65.64 & 67.98 & 45.61 & 49.55 \\ \hline
\end{tabular}
}
\label{tab:Models_Finetuned}
\end{table*}

\subsection{Implementation Details}

Our study was implemented in Python using the PyTorch library on an Nvidia 1080Ti GPU. For fair comparison with Meta-DermDiagnosis \cite{mahajan2020meta}, we matched the base/novel class splits and resized all datasets to 224×224×3. ISIC2018 includes 4 base and 3 novel classes, Derm7pt has 13 base and 5 novel classes, and SD-198 contains 128 base and 70 novel classes. Unlike Meta-DermDiagnosis, we further split SD-198 base classes: those with fewer than 20 samples were set as novel, and those with 20–30 as validation. For consistency, all models were trained with fixed seeds and deterministic settings, ensuring tasks of equal difficulty and order. Hyperparameters such as batch size used during testing were structured similarly for FETL and FEL models, with the only difference being the query set size set to 5 due to data insufficiency in novel classes. Data augmentation techniques are not applied to novel classes. A diverse set of backbone architectures, including convolutional and transformer-based models, was used throughout the experiments, reflecting both conventional and self-supervised transfer learning settings.

The sole distinction between the FEL and FETL models lies in the use of pretrained ImageNet weights for the backbone models in the FETL model. In contrast, the FEL model is trained on backbone models with random initialization. Data augmentation techniques such as RandomResizedCrop, RandomFlip, and ColorJitter are employed. In this framework, our models are trained with a 5-way 5-shot configuration, and all layers of the backbone models are fully opened for training.

Similarly, both DTL, DTLS and DL models utilize ImageNet weights, but the DTL and DTLS model is fine-tuned with the base classes of the datasets. The DL model, serving as a baseline, is chosen to compare performance without any fine-tuning, using ImageNet weights. 
In the DTL and DTLS setups, a standard training approach is utilized instead of episodic training; for instance, in SD-198, 10\% of the base classes are set aside for validation during fine-tuning. DTLS models were pretrained in a self-supervised manner using contrastive loss, and subsequently evaluated by attaching a linear classification head and fine-tuning the models on base classes. The results of this evaluation are presented in Table~\ref{tab:Models_Finetuned}. For models based on Vision Transformer (ViT-Base), fine-tuning was performed using Low-Rank Adaptation (LoRA) to enable efficient parameter updates. For evaluation consistency, all models,including DTL, DTLS, and DL,were tested using a prototypical episodic structure, mimicking few-shot scenarios during inference.

Various augmentation techniques are employed in different training iterations of the DTL model to compare their effects and success rates. While the DTL-base model utilizes RandomResizeCrop and 50\% Horizontal RandomFlip in the SD-198 and ISIC2018 datasets, Resize and 45\% Horizontal and Vertical RandomFlip are used in the Derm7Pt dataset. Subsequently, the DTL-Base section is kept constant, and batch augmentation techniques such as CutMix, MixUp, and ResizeMix are added for comparison. Each added technique is labeled in the result tables as DTL-CutMix or DTL-ResizeMix. Our model named DTL-All-Augment represents a comprehensive model incorporating all three techniques on top of DTL-Base.

\begin{table*}[!]
\centering
\caption{Performance comparison of our models and SOTA methods on the \textbf{SD-198} dataset, reported in F1-scores.}
\begin{tabular}{l|c|cc|cc}
\hline
\multirow{2}{*}{\textbf{Method}} & \multirow{2}{*}{\textbf{Backbone}} & \multicolumn{2}{c|}{\textbf{2 Way}}                                  & \multicolumn{2}{c}{\textbf{5 Way}} \\ \cline{3-6} 
                                 &                                    & \textbf{1 Shot}               & \multicolumn{1}{c|}{\textbf{5 Shot}} & \textbf{1 Shot}  & \textbf{5 Shot} \\ \hline

PCN \cite{prabhu2019few}     & \multirow{2}{*}{Conv4}               & 70.78±1.61 & 85.87±1.12 & 45.59±1.03 & 65.70±1.02 \\
SCAN     &                                      & 78.00±1.51 & 91.01±0.90 & 55.60±1.07 & 75.65±0.87 \\ \hline
SCAN \cite{li2023dynamic}   &\multirow{1}{*}{Conv6}                                     & 77.64±1.50 & 88.28±1.03 & 54.07±1.24 & 74.73±0.92 \\ \hline
NCA \cite{wu2018improving}      & \multirow{8}{*}{WRN-28-10}          & 71.27±1.50 & 84.23±1.19 & 45.91±1.08 & 62.83±1.01 \\
Baseline \cite{chen2019closer}   &                                   & 76.64±1.56 & 89.66±0.97 & 52.54±1.11 & 74.71±0.96 \\

S2M2\_R \cite{mangla2020charting}    &                                   & 77.15±1.59 & 90.97±0.89 & 55.49±1.13 & 78.17±0.84 \\
NegMargin \cite{liu2020negative}  &                                   & 77.98±1.45 & 90.65±0.92 & 56.04±1.14 & 77.75±0.87 \\
PT+NCM \cite{hu2021leveraging}    &                                    & 78.86±1.47 & 90.90±0.93 & 56.91±1.11 & 78.12±0.88 \\
PEM\textsubscript{b}E-\textup{NCM} \cite{wu2018improving} &            & 78.70±1.49 & 90.94±0.95 & 57.42±1.11 & 78.78±0.90 \\
EASY \cite{bendou2022easy}    &                                      & 79.44±1.51 & 91.43±0.96 & 57.77±1.12 & 79.53±0.89 \\
SCAN \cite{li2023dynamic}      &                                     & 81.21±1.46 & 92.08±0.85 & 58.75±1.14 & 81.43±0.77 \\ \hline
DTL+Aug(Ours) & ResNet50 & 75.04±0.38 & 94.11±0.62 & 55.93±0.43 & 86.55±0.71 \\ \hline
DTL+Aug(Ours) & ViT(Base) & 81.76±0.65 & 92.10±0.45 & 64.00±0.37 & 85.10±0.59 \\ \hline
DTL+Aug(Ours) & DenseNet121 & \underline{82.29±0.47} & \underline{95.46±0.36} & \underline{65.54±0.66} & \underline{88.76±0.42} \\ \hline
DTL+Aug(Ours) & MobileNetV2 & \textbf{84.63±0.51} & \textbf{95.29±0.27} & \textbf{67.94±0.40} & \textbf{88.49±0.25} \\ \hline

\end{tabular}
\smallskip 
\footnotesize 

\centering

\textbf{Note:} The results of the SOTA models are taken from the SCAN \cite{li2023dynamic}.

\label{tab:Sota_SD198}

\end{table*}

\subsection{Experimental Analysis}
\label{sec:ExpAnalysis}
In order to evaluate our framework and assess the effectiveness of the methods, we conducted experiments using five few-shot configurations: 2-Way 1-Shot (2W1S), 2-Way 5-Shot (2W5S), 2-Way 10-Shot (2W10S), 5-Way 1-Shot (5W1S), and 5-Way 5-Shot (5W5S). These tests are conducted using three datasets: SD-198, Derm7pt, and ISIC2018. Model names in the results Table-\ref{tab:Models_Finetuned} follow a consistent notation where, for example, \textit{ResNet50(SimCLR)} refers to a ResNet50 backbone that was first pretrained using the SimCLR self-supervised method and then fine-tuned on base classes for classification. The “+Aug” suffix indicates the application of batch-level data augmentation techniques such as MixUp, CutMix, or ResizeMix during fine-tuning. The consolidated results are presented in Tables \ref{tab:Models_Backboned} and \ref{tab:Models_Finetuned}. Table \ref{tab:Models_Backboned} summarizes the performances of backbone models without fine-tuning, while Table \ref{tab:Models_Finetuned} presents the results of models fine-tuned under different training strategies, including FEL, FETL, DTL, DTLS, and baseline DL. Hence  both the inherent capability of pretrained encoders and the improvements obtained through fine-tuning processes under few-shot learning settings is obtained.

Moreover, we executed additional experiments based on the parameters specified in Tables \ref{tab:Sota_SD198}, \ref{tab:Sota_ISIC2018}, and \ref{tab:Sota_Derm7pt}. These experiments are designed to compare our model's performance against benchmarks set in prior research, such as SCAN \cite{li2023dynamic}, MetaMed \cite{singh2021metamed}, and PFEMed \cite{dai2023pfemed}, using the datasets referred to in Section \cite{sd198dataset,derm7ptdataset,isic2018dataset}. In all these tables, the highest accuracy values are highlighted in bold, and the second highest results are underlined. 

The SD-198 dataset contains 198 classes and consists only of clinical images. In contrast, the Derm7pt dataset has 18 classes with a mix of clinical and dermoscopic images. The ISIC2018 dataset, with its 7 classes, mainly features dermoscopic images. Although each of these three datasets shows a long-tail distribution, their unique features affect how models perform. 
In addition to dataset characteristics, model capacity in terms of parameter count and training computation cost (FLOPs) plays a significant role for few-shot classification performance. As detailed in Table~\ref{tab:Models_Backboned}, the tested models vary widely in size, ranging from lightweight architectures such as MobileNetV2 (3.5M parameters, 0.32G FLOPs) to heavyweight models like ViT-Base (86.6M parameters, 17.6G FLOPs). In our experiments, while compact models such as MobileNetV2 and DenseNet121 generally provided stable performance across few-shot setups due to their simplicity and lower risk of overfitting, larger models like ViT-Base demonstrated notably strong performance when appropriately fine-tuned on Derm7pt. 
These findings highlight that smaller capacity model alone is not a disadvantage in few-shot skin disease classification; rather, the effectiveness of large models depends on the availability of diverse and well-aligned data for adaptation.

Following the evaluation of Table~\ref{tab:Models_Backboned}, several backbone architectures, including MobileNetV2, DenseNet121, ViT-Base, and ResNet50, were selected for fine-tuning experiments due to their consistent performance across different datasets and shot configurations. These models demonstrated a favorable balance between model complexity and few-shot generalization capability. To further investigate the impact of contrastive self-supervised pretraining, we included ResNet50 models pretrained with methods such as SimCLR, SimSiam, and MoCo-v3 in our evaluations.

We have also included foundation models such as BioViL and MedCLIP in our evaluation by utilizing their vision encoders directly for few-shot classification. These models were originally pretrained on chest X-ray datasets using paired image–text data. When applied to our skin disease datasets, they performed below ImageNet-pretrained counterparts. This is likely due to a domain mismatch, as these models were trained on radiographic images with associated clinical text, whereas our datasets consist of dermoscopic and clinical skin images without accompanying text. Additionally, since we did not perform further fine-tuning due to lack of domain-specific textual data, their representations could not be adapted to the target domain, limiting their effectiveness in few-shot settings.

First, across all datasets, FETL models consistently outperformed FEL models, particularly in lower-shot settings (e.g., 2W1S, 5W1S), highlighting the critical advantage of using pretrained ImageNet weights in few-shot scenarios. As the number of shots increased, transfer learning-based models, especially DTL and DTL, demonstrated significant gains over both FEL and FETL models. For instance, in the SD-198 dataset under the 5W5S setup, MobileNetV2 achieved 83.22\% accuracy with FETL, while DTL with augmentation (DTL+Aug) reached 88.49\%, clearly illustrating the impact of full fine-tuning and advanced data augmentation strategies. Similarly, DenseNet121 models showed a comparable trend: 84.02\% with FETL and 88.76\% with DTL+Aug. The behaviour of the models are similar in Derm7pt dataset. The reason behind this phenomenon lies in the adaptability of episodic learning, which performs better in scenarios with fewer shots due to the large number of classes. Additionally, clinical images inherently encapsulate various differences, making models trained in an episodic manner more inclined to tolerate these diversities. As the number of shots increases, models trained non-episodically become more successful in classifications.

When observing Derm7pt results, ViT-Base models adapted with LoRA fine-tuning exhibited outstanding performance, achieving over 93\% accuracy in low-shot settings (2W5S and 5W5S). This suggests that large-capacity models, when fine-tuned properly on diverse datasets, can effectively overcome the challenges posed by few-shot classification.

The evaluation results presented in Table~\ref{tab:Models_Finetuned} demonstrate that among the contrastive pretraining strategies tested, SimCLR consistently yields the most favorable performance across all datasets when combined with augmentation and supervised fine-tuning (DTLS). For example, in the Derm7pt dataset, ResNet50(SimCLR)+Aug outperforms SimSiam and MoCo-v3 in nearly all configurations, particularly in 2W5S and 5W5S settings, where it achieves 80.58\% and 63.26\% accuracy, respectively. On ISIC2018, which has fewer classes and a more homogeneous modality, all contrastive methods perform comparably, but SimCLR still maintains a slight edge. These results suggest that SimCLR-based self-supervised pretraining provides a more transferable representation for downstream few-shot tasks, especially when followed by targeted fine-tuning with data augmentation. Overall, the choice of contrastive method can significantly affect performance, highlighting the importance of selecting an appropriate pretraining strategy for different dataset characteristics.

\begin{table*}[!]
\centering
\caption{Performance comparison of our models and SOTA methods on the \textbf{ISIC2018} dataset, reported in accuracy (\%).}

\begin{tabular}{l|c|ccc|ccc}
\hline
\multirow{2}{*}{\textbf{Method}} & \multirow{2}{*}{\textbf{Backbone}} & \multicolumn{3}{c|}{\textbf{2 Way}}                                                                & \multicolumn{3}{c}{\textbf{3 Way}}                                                                \\ \cline{3-8} 
                                 &                                     & \textbf{3 Shot}      & \textbf{5 Shot} & \textbf{10 Shot} & \textbf{3 Shot}      & \textbf{5 Shot} & \textbf{10 Shot} \\ \hline                             
                                 
Meta-DermDiagnosis \cite{mahajan2020meta}     & \multirow{1}{*}{Conv6}       & 64.50           & 73.50           & 79.70            & -               & -               & -                \\ \hline
MetaMed - Transf. Learn. \cite{singh2021metamed}  & \multirow{5}{*}{Conv4}       & 66.88           & 73.88           & 81.37            & 54.83           & 59.33           & 69.75            \\
MetaMed - Normal Aug. \cite{singh2021metamed}     &                              & 72.75           & 75.62           & 81.37            & 54.83           & 59.33           & 69.75            \\
MetaMed - CutOut \cite{singh2021metamed}          &                              & 70.37           & 77.62           & 81.87            & 55.50           & 65.41           & 69.75            \\
MetaMed - MixUp \cite{singh2021metamed}           &                              & 75.37           & 78.25           & 84.25            & 58.50           & 61.25           & 71.00            \\
MetaMed - CutMix \cite{singh2021metamed}          &                              & 73.25           & 76.87           & 80.62            & 58.66           & 61.50           & 66.50            \\ \hline
PT-MAP \cite{hu2021leveraging}                    & \multirow{5}{*}{WRN}         & 68.15           & 70.87           & 74.19            & 53.17           & 55.61           & 59.57            \\
Baseline+ \cite{chen2019closer}                   &                              & 64.77           & 70.27           & 74.67            & 53.20           & 54.16           & 57.87            \\
NegMargin \cite{liu2020negative}                  &                              & 71.33           & 72.67           & 75.17            & 60.69           & 57.58           & 63.04            \\
Baseline \cite{chen2019closer}                    &                              & 68.77           & 71.03           & 76.97            & 56.80           & 59.20           & 65.22            \\ 
PFEMed \cite{dai2023pfemed}                       &                              & \textbf{81.69}  & \textbf{83.87}  & 85.14            & \textbf{66.94}  & 69.78           & 73.81            \\ \hline
DTL+Aug (Ours)          &   ResNet50        & 73.57  & 78.28  & 83.47      & 59.79  & 65.86  & 72.70   \\  \hline
DTL+Aug (Ours)          &   DenseNet121     & 62.42  & 65.06  & 67.85      & 45.74  & 48.66  & 52.34   \\ \hline
DTL+Aug (Ours)          &   ViT(Base)       & 62.85  & 65.64  & 67.98      & 47.22  & 49.55  & 53.25   \\ \hline

DTL+Base (Ours)            & \multirow{5}{*}{MobileNetV2} & 77.26           & 81.63           & 85.18            & 63.98           & 69.35           & 75.01            \\ 
DTL+CutMix (Ours)          &                              & 77.71           & 81.79           & 85.97            & 64.37           & 69.86           & 75.73            \\ 
DTL+MixUp (Ours)           &                              & \underline{79.02} & 82.95           & 86.40            & \underline{66.84} & \underline{71.15} & 76.48            \\ 
DTL+ResizeMix (Ours)       &                              & 78.02           & 82.75           & \underline{86.69} & 65.56           & 70.87           & \textbf{76.97}   \\ 
DTL+AllAug (Ours)     &                              & 78.96           & \underline{83.21} & \textbf{86.83}  & 66.12           & \textbf{71.28}  & \underline{76.94}   \\ \hline

\end{tabular}

\smallskip 
\footnotesize 

\centering

\textbf{Note:}The results of the SOTA models are taken from the PFEMed \cite{dai2023pfemed}.

\label{tab:Sota_ISIC2018}
\end{table*}

Since the ISIC2018 dataset comprises dermoscopic data, the discriminative power of domain-specific features becomes more crucial. Hence, transfer learning-based DTL and DL models outperform FETL and FEL models. For instance, for MobileNetV2, the accuracy rates are 58.49\% for 2W1S with FETL and 64.83\% with DTL, and 68.82\% for 2W10S with FETL and 85.18\% with DTL (Table \ref{tab:Models_Finetuned}).  Episodic learning-based methods tend to remain superficial in training.

In all three datasets we studied, transfer learning methods generally perform better than episodic few-shot training in learning distinct features, except in cases where there is only one example provided. This observation suggests that even though episodic few-shot training is a more intricate approach, it may not be as effective as transfer learning in most scenarios. The exception is when extremely limited data is available, which is when episodic few-shot training can be valuable. This finding is consistent with recent research (\cite{laenen2021episodes}) that questions the practicality of complex few-shot training methods.

Our proposed DTL approach exhibited promising performance on datasets containing clinical data, particularly SD-198. However, its performance was relatively lower on dermoscopic datasets such as ISIC2018 and, to some extent, Derm7pt. As illustrated in Table~\ref{tab:Models_Finetuned}, DenseNet121-based models generally performed worse than MobileNetV2-based counterparts on ISIC2018, especially under low-shot conditions such as 2W1S and 3W1S. For example, under the 2W1S setup, DenseNet121 DTL achieved 67.51\% accuracy, while MobileNetV2 DTL reached 69.5\%. This gap can be attributed to DenseNet121's tendency to overfit in low-data regimes. Moreover, ISIC2018’s structure (with only four base and three novel classes) offers limited variation for deeper architectures to generalize effectively. On Derm7pt, both MobileNetV2 and DenseNet121 showed more comparable results, with DenseNet121+Aug performing competitively in 5W5S. Nonetheless, across both datasets, neither model outperformed recent state-of-the-art methods such as SCAN or PFEMed. These results emphasize the need for improved techniques that specifically address the unique challenges of dermoscopic image classification, including fine-grained visual patterns and modality-specific artifacts. Notably, the ViT(Base)+LoRA+Aug model achieved the highest accuracy across several configurations on Derm7pt (e.g., 2W5S and 5W5S).

Our experiments further demonstrate that the transfer learning strategy in the FETL,DTL and DTLS methods effectively incorporates advances from natural image domains into skin disease classification tasks. Datasets include images with hair, ruler marks, varying body regions, and differences in skin tone. These artifacts may introduce instability in the models. However, leveraging data augmentation techniques like Flip and Resize, along with mix-based methods such as MixUp, CutMix, and ResizeMix, not only boosted model accuracy but also enhanced model robustness. A more detailed analysis and comprehensive discussion of these findings can be found in Section~\ref{sec:additional}.

One limitation of our study is the absence of evaluation on independent clinical datasets, due to limited access beyond public benchmarks. While our models generalize well on standard datasets, future work will focus on external clinical validation to better assess real-world applicability.

\subsection{Model Explainability}
\label{sec:model_explainablity}

\begin{figure}[!t]
	\centering
	\includegraphics[width=0.8\textwidth]{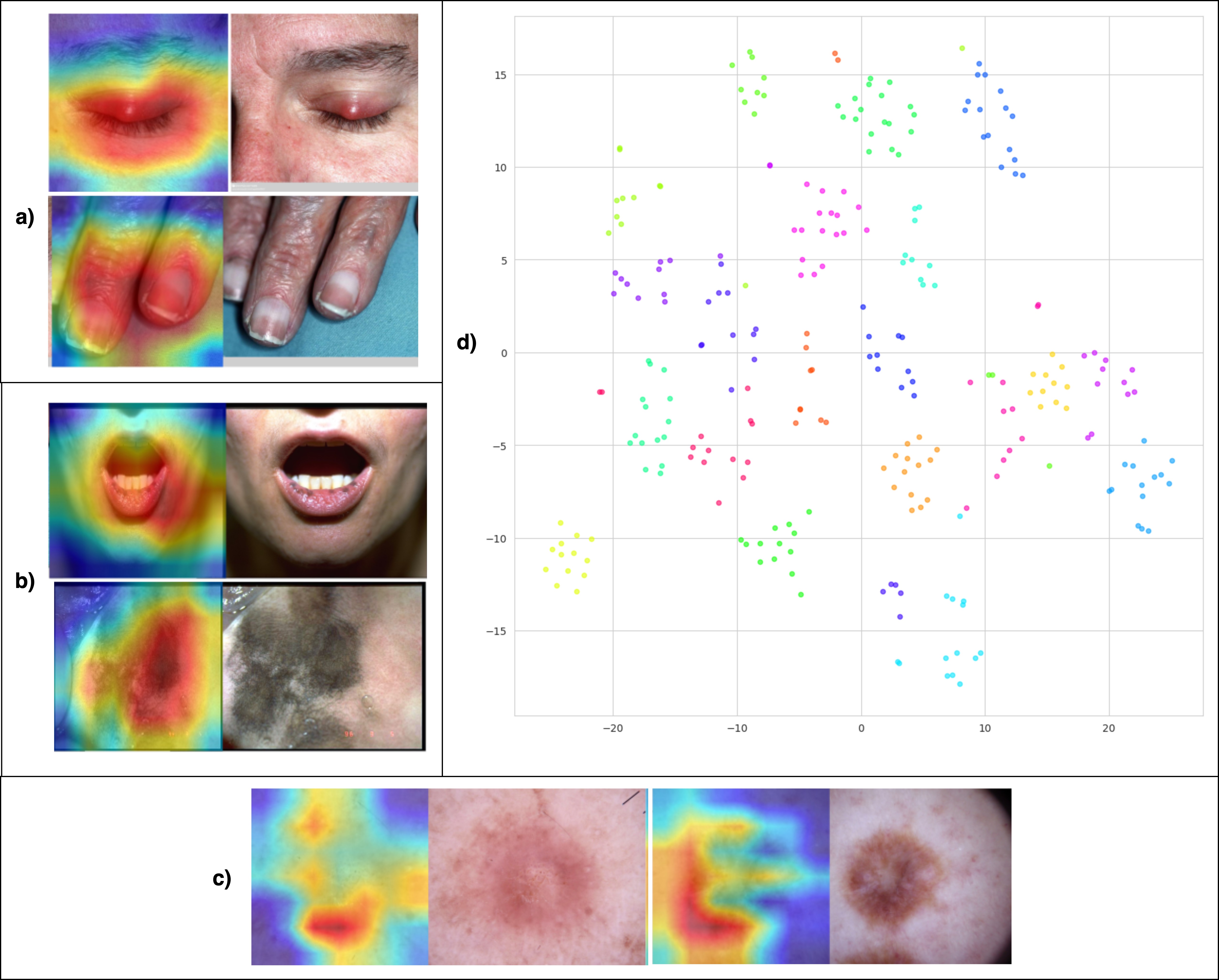}	
    \caption{
    (a) Grad-CAM on SD-198, 
    (b) Grad-CAM on Derm7pt, 
    (c) Grad-CAM on ISIC2018, 
    (d) t-SNE of 20 randomly selected SD-198 novel classes.
    }
	\label{gradcam}
\end{figure}

\begin{figure*}[ht]
	\centering 
	\includegraphics[width=\textwidth]{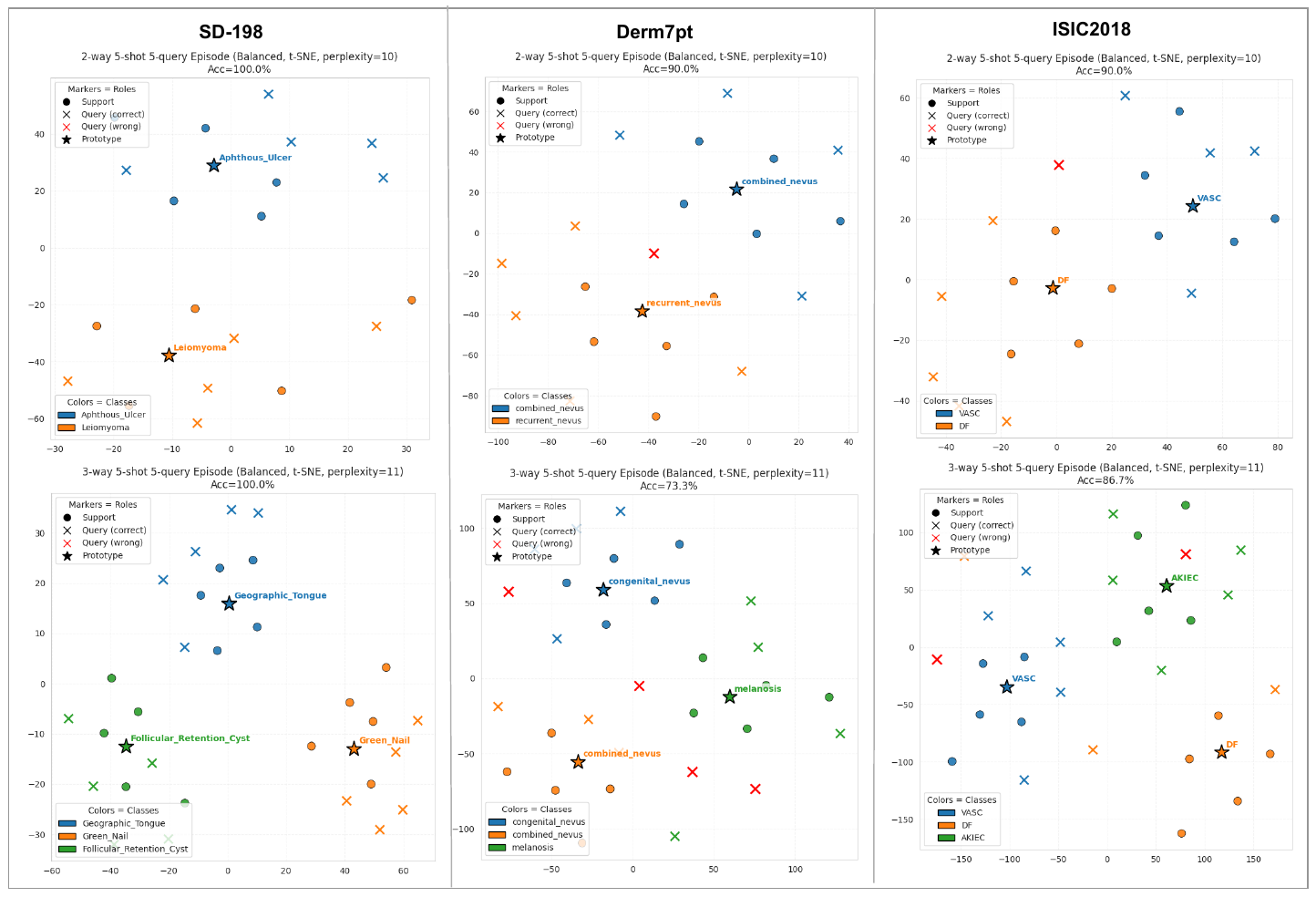}	
	\caption{Prototype visualization with t-SNE on ISIC2018, Derm7Pt, and SD-198. Please refer to the digital version for better visibility.}

	\label{prototype_tsne}
\end{figure*}

\begin{figure}[!t]
	\centering
	\includegraphics[width=\textwidth]{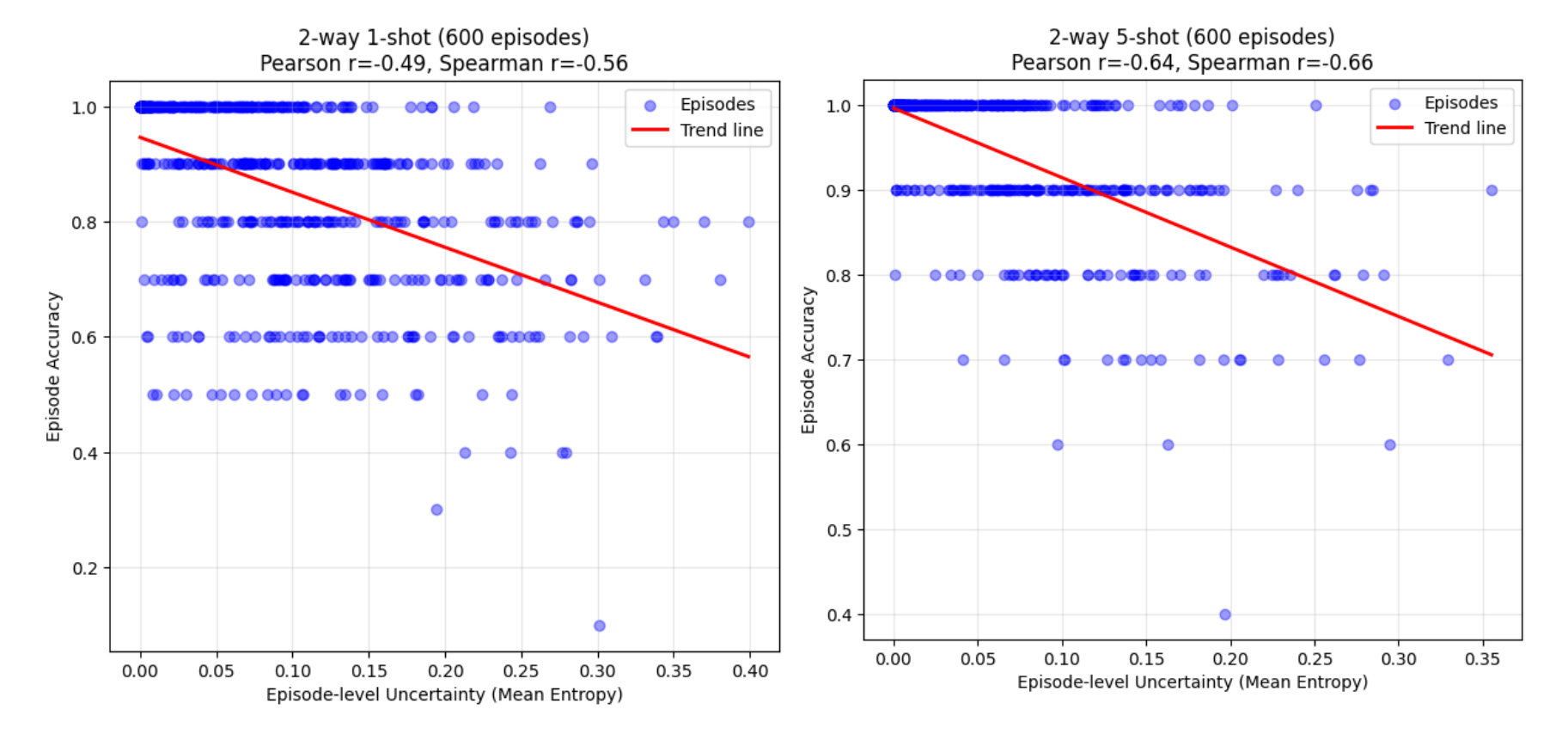}	
    \caption{Episode-level predictive uncertainty (mean entropy) versus episode accuracy for the SD-198 dataset with the MobileNetV2 model using Monte Carlo dropout.}

	\label{entropy}
\end{figure}

To improve interpretability, we applied three common explainability techniques: prototype visualization with t-SNE \cite{maaten2008visualizing}, Grad-CAM \cite{selvaraju2017grad}, and predictive uncertainty estimation using Monte Carlo dropout \cite{gal2016dropout}. All results were obtained using the DTL+Aug MobileNetV2 model across all datasets.

Grad-CAM maps (Figure~\ref{gradcam}), which are obtained from the last convolutional layer of the embedding extractor, show that the model generally attends to lesion regions in clinical datasets (SD-198 and Derm7pt), while attention in dermoscopic images (ISIC2018) is more scattered, which helps explain some misclassifications. Moreover, the t-SNE visualization of 20 randomly selected SD-198 classes in Figure~\ref{gradcam}(d) demonstrates that the base-trained model, despite never observing novel classes, is still able to form reasonably separable clusters in the embedding space.

In addition, we performed t-SNE visualization during testing using prototypical embeddings, particularly focusing on 2-way and 3-way episodes. Across all three datasets, distinct clustering is observed in the embedding space; however, clustering patterns are clearer in the clinical datasets (SD-198 and Derm7pt), while ISIC2018 remains less separable due to its higher visual similarity among classes (Figure~\ref{prototype_tsne}).

Finally, the uncertainty analysis (Figure~\ref{entropy}) revealed complementary insights at two levels. At the episode level, predictive entropy showed a negative correlation with accuracy (Pearson $r=-0.49$ for 2-way 1-shot and $r=-0.64$ for 2-way 5-shot). At the query level, misclassified samples tended to have higher entropy values, indicating that uncertainty is a useful signal for identifying unreliable predictions. Increasing the number of support samples from 1 to 5 reduced overall uncertainty and strengthened its correlation with accuracy, showing that predictions became more stable and entropy served as a more reliable error indicator. In addition to the overall decrease, entropy distributions became narrower with more support samples, indicating more stable predictions across episodes.

\subsection{Comparison with Current State-of-the-Art}
\label{sec:SOTA}
We also included a thorough analysis aimed at understanding how different skin disease datasets, each with unique features and conditions, influence the training process. We arranged the datasets in our research in a manner similar to SCAN \cite{li2023dynamic} and PFEMed \cite{dai2023pfemed} studies  for a fair comparison of the model performances. Specifically, we examined the performance benchmarks set by the SCAN model on the SD-198 and Derm7pt datasets, and by the PFEMed model on the ISIC2018 dataset.

In our evaluation on the SD-198 dataset, we enhanced our Deep Transfer Learning (DTL) models using various data augmentation techniques and compared them against recent state-of-the-art methods, including SCAN, EASY, PT+NCM, and others (Table~\ref{tab:Sota_SD198}). Unlike the SCAN model, which uses a WRN-28-10 backbone with an unsupervised clustering branch, our models leverage lighter backbones such as MobileNetV2 and ViT, combined with ImageNet pretraining and multiple augmentation strategies (e.g., MixUp, CutMix, ResizeMix). Notably, our MobileNetV2-based DTL+AllAug model outperformed all prior methods in every configuration, achieving the highest F1-scores across all 2-Way and 5-Way shot settings. This indicates that even with a simpler architecture and traditional transfer learning, effective data augmentation can result in state-of-the-art performance.

On the ISIC2018 dataset, we compared our results with benchmarks established by PFEMed, MetaMed, PT-MAP, and others (Table~\ref{tab:Sota_ISIC2018}). PFEMed, which incorporates a dual-encoder architecture with variational modeling, achieved strong results, particularly in low-shot 3-Way setups. However, our MobileNetV2 DTL+AllAug model achieved the highest accuracy in several 2-Way and 3-Way configurations, including 2W10S (86.83\%) and 3W10S (76.94\%). Furthermore, our ViT-Base-based model also showed competitive results, confirming that transformer backbones can generalize well in few-shot medical image settings when properly fine-tuned. These findings suggest that our approach can match or even surpass more complex meta-learning-based methods, particularly when sufficient training shots are available.

We further compared our models against state-of-the-art methods on the Derm7pt dataset. As shown in Table~\ref{tab:Sota_Derm7pt}, our ViT(Base)+Aug DTL model achieved the highest accuracies in both 2-Way 1-Shot (86.35\%) and 2-Way 5-Shot (93.68\%) settings, outperforming strong baselines such as SCAN and PFEMed. In addition, MobileNetV2 and DenseNet121-based DTL models achieved competitive performance, especially when paired with augmentation techniques such as MixUp and CutMix. These results reinforce the notion that a well-designed transfer learning framework, which combines pretrained weights, architectural diversity, and batch-level augmentation, can effectively tackle the unique challenges posed by long-tailed dermatological datasets, without relying on meta-learning or external pretraining data.

\begin{table}[!]
 
\centering
\caption{Performance of our models and SOTA methods on the \textbf{Derm7pt} dataset, reported in Accuracy (\%).}

\begin{tabular}{l|c|cc}
\hline
\multirow{2}{*}{\textbf{Method}} & \multirow{2}{*}{\textbf{Backbone}} & \multicolumn{2}{c}{\textbf{2 Way}}                         \\ \cline{3-4} 
                                 &                                    & \textbf{1 Shot}      & \textbf{5 Shot} \\ \hline

PCN \cite{prabhu2019few}                      & \multirow{2}{*}{Conv4}       & 59.98±1.28           & 70.62±1.3        \\
SCAN \cite{li2023dynamic}                    &                              & 61.42±1.49           & 72.58±1.28      \\ \hline
Meta-DermDiagnosis \cite{mahajan2020meta}    & \multirow{2}{*}{Conv6}       & 61.8                 & 76.9            \\
SCAN \cite{li2023dynamic}                    &                              & 62.80±1.34           & 76.65±1.21      \\ \hline
NCA \cite{wu2018improving}                   & \multirow{9}{*}{WRN-28-10}   & 56.32±1.29           & 67.18±1.15      \\
Baseline \cite{chen2019closer}              &                              & 59.43±1.34           & 74.28±1.14      \\
S2M2\_R \cite{mangla2020charting}           &                              & 61.37±1.33           & 79.83±1.34      \\
NegMargin \cite{liu2020negative}           &                              & 58.00±1.44           & 70.12±1.30      \\
PT+NCM \cite{hu2021leveraging}              &                              & 60.92±1.68           & 74.33±1.48      \\
PEMb E\_NCM \cite{wu2018improving}          &                              & 60.40±1.72           & 72.63±1.48      \\
EASY \cite{bendou2022easy}                  &                              & 61.02±1.67           & 75.98±1.41      \\
SCAN \cite{li2023dynamic}                   &                              & 66.75±1.35           & 82.57±1.13 \\
PFEMed \cite{dai2023pfemed}                 &                              & 71.15    & 80.27           \\ \hline

DTL+Aug (Ours) & ResNet50 & 61.62±0.53 & 78.86±0.47 \\ \hline
DTLS+Aug (Ours) & ResNet50(SimCLR) & 63.91±0.38 & 80.58±0.44 \\ \hline
DTL+Aug (Ours) & MobileNetV2 & 60.37±0.69 & 77.41±0.41 \\ \hline
DTL+Aug (Ours) & DenseNet121 & 62.56±0.40 & 81.00±0.39 \\ \hline
DL (Ours) & MoCo-v3(ViT) & \textbf{88.40±0.55} & \underline{92.18±0.63} \\ \hline
DTL+Aug (Ours) & ViT-Base & \underline{86.35±0.66} & \textbf{93.68±0.48} \\ \hline

\end{tabular}

\smallskip 
\footnotesize 

\centering

\textbf{Note:}The results of the SOTA models are taken from the SCAN \cite{li2023dynamic}. 

\label{tab:Sota_Derm7pt}
\end{table}

\section{Additional Analysis}
\label{sec:additional}

We conducted additional analysis to examine how various augmentation strategies affect the performance of our DTL models on three benchmark datasets: SD-198, Derm7pt, and ISIC2018. We employed the following strategies: (1) Base (random horizontal and vertical flips at 45°), (2) CutMix, (3) MixUp, (4) ResizeMix, and (5) AllAug, which combines all the aforementioned methods on top of the Base augmentation. Detailed results for the SD-198, Derm7pt, and ISIC2018 datasets are presented in Tables~\ref{tab:Augmentation_Detail}.

We previously mentioned in Section \ref{sec:ExpAnalysis} that artifacts in skin disease datasets compromise model stability, yet mix-based augmentation techniques effectively address these issues. Notably, the SD-198 dataset contains clinical images with varying artifacts (i.e. high scale variances within classes) due to different camera setups. Since ResizeMix resizes one image and overlays it onto another, it enhances the model’s resilience to such variations. Consequently, ResizeMix outperformed CutMix, MixUp, and AllAug by approximately 3\% on the SD-198 dataset. In contrast, dermoscopy datasets, such as Derm7pt and ISIC2018, typically involve specialized imaging devices that capture deeper layers of the skin at uniform depths, minimizing distance-related variability and other artifacts. As a result, ResizeMix demonstrated similar performance to CutMix and MixUp on these two datasets.

Finally, we investigated how the architectural characteristics of MobileNetV2 and DenseNet121 influence the outcomes of these augmentations. Both backbones achieved comparable performance on the SD-198 and Derm7pt datasets. However, as discussed in the Discussion section, DenseNet121 suffered from overfitting on the ISIC2018 dataset and consequently failed to deliver robust results.


\begin{table}[]
\centering
\caption{Accuracy (\%) comparison of models and augmentation techniques on the \textbf{SD-198}, \textbf{Derm7Pt}, and \textbf{ISIC2018} datasets.}

\begin{tabular}{llllll}
\hline
\textbf{\begin{tabular}[c]{@{}l@{}}Dataset\\ Model\end{tabular}} & \textbf{Method} & \textbf{2W-1S} & \textbf{2W-5S} & \textbf{5W-1S} & \textbf{5W-5S} \\ \hline
\multirow{5}{*}{\textbf{\begin{tabular}[c]{@{}l@{}}SD-198\\ MobileNetV2\end{tabular}}} & Base & 82.42 & 94.40 & 64.06 & 86.67 \\
 & Base+CutMix & 83.00 & 94.77 & 65.48 & 87.63 \\
 & Base+Mixup & 82.29 & 94.48 & 64.56 & 87.22 \\
 & Base+ResizeMix & \textbf{84.63} & \textbf{95.29} & \textbf{67.94} & \textbf{88.49} \\
 & Base+AllAug & 83.31 & 95.02 & 65.99 & 88.04 \\ \hline
\multirow{5}{*}{\textbf{\begin{tabular}[c]{@{}l@{}}SD-198\\ DenseNet121\end{tabular}}} & Base & 81.21 & 94.56 & 62.61 & 86.95 \\
 & Base+CutMix & 77.61 & 93.95 & 58.23 & 86.44 \\
 & Base+Mixup & 78.88 & 94.40 & 61.21 & 87.10 \\
 & Base+ResizeMix & \textbf{82.29} & \textbf{95.46} & \textbf{65.54} & \textbf{88.76} \\
 & Base+AllAug & 80.80 & 95.27 & 63.59 & 88.72 \\ \hline
\multirow{5}{*}{\textbf{\begin{tabular}[c]{@{}l@{}}Derm7pt\\ MobileNetV2\end{tabular}}} & Base & 59.65 & 75.41 & 31.82 & 56.10 \\
 & Base+CutMix & \textbf{60.97} & 77.06 & \textbf{33.84} & 59.02 \\
 & Base+Mixup & 60.65 & 76.92 & 33.21 & 58.57 \\
 & Base+ResizeMix & 60.37 & \textbf{77.41} & 33.54 & \textbf{59.45} \\
 & Base+AllAug & 60.53 & 77.10 & 33.37 & 59.05 \\ \hline
\multirow{5}{*}{\textbf{\begin{tabular}[c]{@{}l@{}}Derm7pt\\ DenseNet121\end{tabular}}} & Base & 60.00 & 77.25 & 32.04 & 57.68 \\
 & Base+CutMix & 61.57 & 78.65 & \textbf{35.46} & 60.58 \\
 & Base+Mixup & \textbf{62.56} & \textbf{81.00} & 34.75 & \textbf{61.91} \\
 & Base+ResizeMix & 61.09 & 79.18 & 33.89 & 61.22 \\
 & Base+AllAug & 60.88 & 78.30 & 33.19 & 59.95 \\ \hline
\textbf{\begin{tabular}[c]{@{}l@{}}Dataset\\ Model\end{tabular}} & \textbf{Method} & \textbf{2W-1S} & \textbf{2W-5S} & \textbf{3W-1S} & \textbf{3W-5S} \\ \hline
\multirow{5}{*}{\textbf{\begin{tabular}[c]{@{}l@{}}ISIC2018\\ MobileNetV2\end{tabular}}} & Base & 64.83 & 81.63 & 49.51 & 69.35 \\
 & Base+CutMix & 64.94 & 81.79 & 49.93 & 69.86 \\
 & Base+Mixup & \textbf{66.34} & 82.95 & \textbf{51.66} & 71.15 \\
 & Base+ResizeMix & 64.99 & 82.75 & 50.34 & 70,87 \\
 & Base+AllAug & 65.55 & \textbf{83.21} & 50.40 & \textbf{71.28} \\ \hline
\multirow{5}{*}{\textbf{\begin{tabular}[c]{@{}l@{}}ISIC2018\\ DenseNet121\end{tabular}}} & Base & 55.55 & 64.69 & 38.63 & 46.47 \\
 & Base+CutMix & 54.45 & 60.78 & 38.63 & 44.32 \\
 & Base+Mixup & 55.20 & 64.42 & 39.72 & \textbf{48.98} \\
 & Base+ResizeMix & 55.90 & 63.57 & \textbf{39.85} & 48.27 \\
 & Base+AllAug & \textbf{56.67} & \textbf{65.06} & 39.81 & 48.66 \\ \hline
\end{tabular}

\label{tab:Augmentation_Detail}
\end{table}

\section{Conclusion and Future Work}

In this study, we explored the effectiveness of supervised and self-supervised transfer learning strategies combined with few-shot learning to classify rare skin diseases under long-tail data distributions. While prior work has highlighted the challenges of imbalance and data scarcity in this domain, there remains limited research evaluating the combined impact of transfer learning and few-shot methods in a unified setting. To address this gap, we systematically evaluated five training paradigms (FETL, FEL, DTL, DTLS, and DL) across three benchmark skin image datasets. Our findings demonstrate that DTL models, particularly those using MobileNetV2 backbones and enhanced with MixUp, CutMix, and ResizeMix, consistently outperformed alternative approaches. In addition, transformer-based models such as ViT-Base, fine-tuned via LoRA, achieved state-of-the-art performance on the Derm7pt dataset.

These results reaffirm the strength of conventional transfer learning in few-shot medical imaging tasks and highlight the value of appropriate data augmentation and backbone selection. While episodic and self-supervised methods showed advantages in specific low-shot scenarios, their performance was generally less stable compared to supervised DTL approaches. Additionally, contrastive self-supervised pretraining strategies such as SimCLR and MoCo-v3 were also applied and yielded competitive results in selected scenarios. The datasets used in this study only contained class labels and did not provide additional clinical context such as lesion descriptions, anatomical site annotations, or physician notes. For this reason, vision–language foundation models (e.g., BioViL, MedCLIP) were evaluated in a zero-shot setting and showed limited transferability due to domain mismatch and the absence of text-based adaptation.

Future research may benefit from exploring domain-specific self-supervised pretraining strategies, lightweight transformer fine-tuning methods, and the integration of multimodal approaches. Recently introduced dermatology datasets with rich textual annotations, such as DermaCon-IN \cite{madarkar2025dermacon}, indicate that the field is moving toward cross-modal resources. Building upon these, future work could explore integration with CLIP-based vision–language models and multimodal LLMs such as GPT-4o or BiomedCLIP to combine lesion images with textual or patient-level context, enabling interactive diagnostic support and zero-shot generalization. Finally, extending the framework to other medical imaging modalities (e.g., X-rays, histopathology, fundus imaging) in parallel with multimodal integration could further support robust classification systems for low-data and rare disease applications.

\bibliographystyle{unsrt}  
\bibliography{references}

\section*{Author Biographies}

\begin{minipage}{0.18\linewidth}
\includegraphics[width=\linewidth]{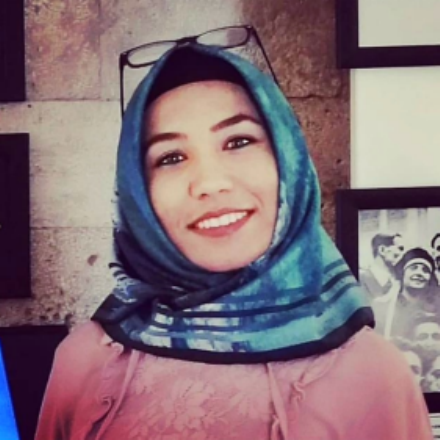}
\end{minipage}\hfill
\begin{minipage}{0.78\linewidth}
\textbf{Zeynep Özdemir} received her B.S. and M.S. degrees in Computer Engineering from Ankara University, Turkey, in 2016 and 2018, respectively, and is currently pursuing her Ph.D. in Computer Engineering at the same institution. 
Since 2018, she has been a Research Assistant in the Department of Computer Engineering at Ankara University. 
Her research interests lie in artificial intelligence for healthcare, with a focus on computer vision, medical image analysis, explainable AI, segmentation, domain adaptation, and rare disease classification. 
Her doctoral research particularly explores few-shot and transfer learning under long-tail distributions for skin disease classification. 
She actively engages with the international MICCAI community through the RISE-MICCAI reading group, further expanding her expertise in medical imaging and computational methods.
\end{minipage}

\vspace{1.5em}

\begin{minipage}{0.18\linewidth}
\includegraphics[width=\linewidth]{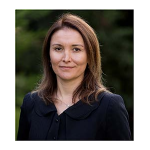}
\end{minipage}\hfill
\begin{minipage}{0.78\linewidth}
\textbf{Dr. Hacer Yalım Keleş} completed her B.Sc., M.Sc., and Ph.D. in Computer Engineering at Middle East Technical University, Turkey, in 2002, 2005, and 2010, respectively. Her Ph.D. thesis was honored with the Thesis of the Year award by the Prof. Dr. Mustafa Parlar Education and Research Foundation in 2010. Between 2000 and 2007, she contributed as a researcher at The Scientific and Technological Research Council of Turkey (TUBITAK), focusing on pattern recognition using multimedia data, including audio and video.

Dr. Keles was an Assistant Professor at the Department of Computer Engineering, Ankara University, from 2013 to 2021, and is currently an Associate Professor at the Department of Computer Engineering, Hacettepe University.

Her research primarily spans computer vision and machine learning, with a focus on learning algorithms for limited data and deep generative models. She has contributed to sign and gesture recognition, generative adversarial networks, image inpainting, and image segmentation domains. Moreover, she collaborates on diverse projects involving aerial and medical images, speech signals, textual, geophysical, and hyperspectral data analysis with her graduate students.
\end{minipage}

\vspace{1.5em}

\begin{minipage}{0.18\linewidth}
\includegraphics[width=\linewidth]{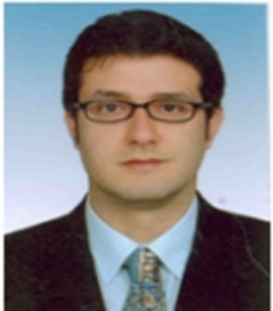}
\end{minipage}\hfill
\begin{minipage}{0.78\linewidth}
\textbf{Dr. Ömer Özgür Tanrıöver} received the B.Sc. degree in computer engineering, the M.Sc. and Ph.D. degrees in information systems from Middle East Technical University, Ankara, Turkey. Previously, he has served as a Certified Information Systems Auditor (CISA) with the Information Management Department, Banking Regulation Agency of Turkey. Currently, he is with Computer Engineering Department of Ankara University as an Associate Professor. His current research interests include applications of machine learning, information systems security and software process.
\end{minipage}

\end{document}